\documentclass[lettersize,journal]{IEEEtran}
\usepackage{amsmath,amsfonts}
\usepackage{array}
\usepackage{textcomp}
\usepackage{stfloats}
\usepackage{url}
\usepackage{verbatim}
\usepackage{graphicx}
\usepackage{color}
\usepackage{subfigure}
\usepackage{booktabs}
\usepackage{multirow}
\usepackage{supertabular}

\newcommand{\ms}[2]{{#1}{\footnotesize $\,\pm${#2}}}
 
\newcommand{\eg}{\textit{e}.\textit{g}.}
\newcommand{\ie}{\textit{i}.\textit{e}.}
\newcommand{\etal}{\textit{et al}.}
\usepackage{array}
\usepackage{cite}
\usepackage[ruled,linesnumbered]{algorithm2e}
\usepackage{bm}
\usepackage[dvipsnames]{xcolor}
\usepackage{colortbl}
\usepackage{xcolor}
\usepackage{bbding}
\definecolor{mypink2}{rgb}{.99,.96,.98}
\definecolor{mypink1}{rgb}{.99,.93,.98}
\definecolor{mypink}{rgb}{.99,.90,.98}
\definecolor{mygray}{rgb}{.95,.95,.95}
\begin{document}

\title{IMWA: Iterative Model Weight Averaging Benefits \\Class-Imbalanced Learning Tasks}

\author{Zitong Huang$^1$, Ze Chen$^2$ , Bowen Dong$^1$, Chaoqi Liang$^1$, Erjin Zhou$^2$, 
Wangmeng Zuo$^1$\textsuperscript{\Envelope} 

\thanks{Zitong Huang, Bowen Dong, Chaoqi Liang and Wangmeng Zuo are with the School of Computer Science and Technology, Harbin Institute of Technology, Harbin 150001, China (e-mail: cswmzuo@gmail.com). 

Ze Chen and Erjin Zhou are with the Megvii Research, Megvii Technology Limited, Beijing, China. 
}}

\markboth{Journal of \LaTeX\ Class Files,~Vol.~14, No.~8, August~2021}%
{Shell \MakeLowercase{\textit{et al.}}: A Sample Article Using IEEEtran.cls for IEEE Journals}


\maketitle

\begin{abstract}
Model Weight Averaging (MWA) is a technique that seeks to enhance model's performance by averaging the weights of multiple trained models. 
This paper first empirically finds that 1) the vanilla MWA can benefit the class-imbalanced learning, and 2) performing model averaging in the early epochs of training yields a greater performance improvement than doing that in later epochs.
Inspired by these two observations, in this paper we propose a novel MWA technique for class-imbalanced learning tasks named Iterative Model Weight Averaging (IMWA).
Specifically, IMWA divides the entire training stage into multiple episodes. 
Within each episode, multiple models are concurrently trained from the same initialized model weight, and subsequently averaged into a singular model. 
Then, the weight of this average model serves as a fresh initialization for the ensuing episode, thus establishing an iterative learning paradigm.
%
Compared to vanilla MWA, IMWA achieves higher performance improvements with the same computational cost.
Moreover, IMWA can further enhance the performance of those methods employing EMA strategy, demonstrating that IMWA and EMA can complement each other.
Extensive experiments on various class-imbalanced learning tasks, \ie, class-imbalanced image classification, semi-supervised class-imbalanced image classification and semi-supervised object detection tasks showcase the effectiveness of our IMWA.
\end{abstract}

\begin{IEEEkeywords}
 Model Weight Averaging, Class-Imbalanced Learning.
\end{IEEEkeywords}

\section{Introduction}\label{sec:intro}

Model Weight Averaging (MWA) aims at merging optimized parameters from different models to obtain model parameters with better prediction accuracy.
Compared to conventional model ensembling methods (\emph{i.e.}, fusing predictions from multiple pretrained models), MWA methods usually obtain similar or even better performance while lead to less computation cost.
Recently, several studies explored the application of MWA in various fields.
For example, Model Soups~\cite{wortsman2022model}   chose candidate weights from dozens of models fine-tuned from pretrained models.
Matena \etal~\cite{matena2022merging} proposed Fisher Merging, which merges various models' functionalities by calculating a weighted average of their respective parameters using the Fisher information.
Other several studies ~\cite{ilharco2022patching,rame2022diverse} illustrated the effects of MWA on continual learning and out-of-distribution scenarios.
%
%
%

%

\begin{figure}[t]

    \begin{center}
    \includegraphics[width=1.0\linewidth]{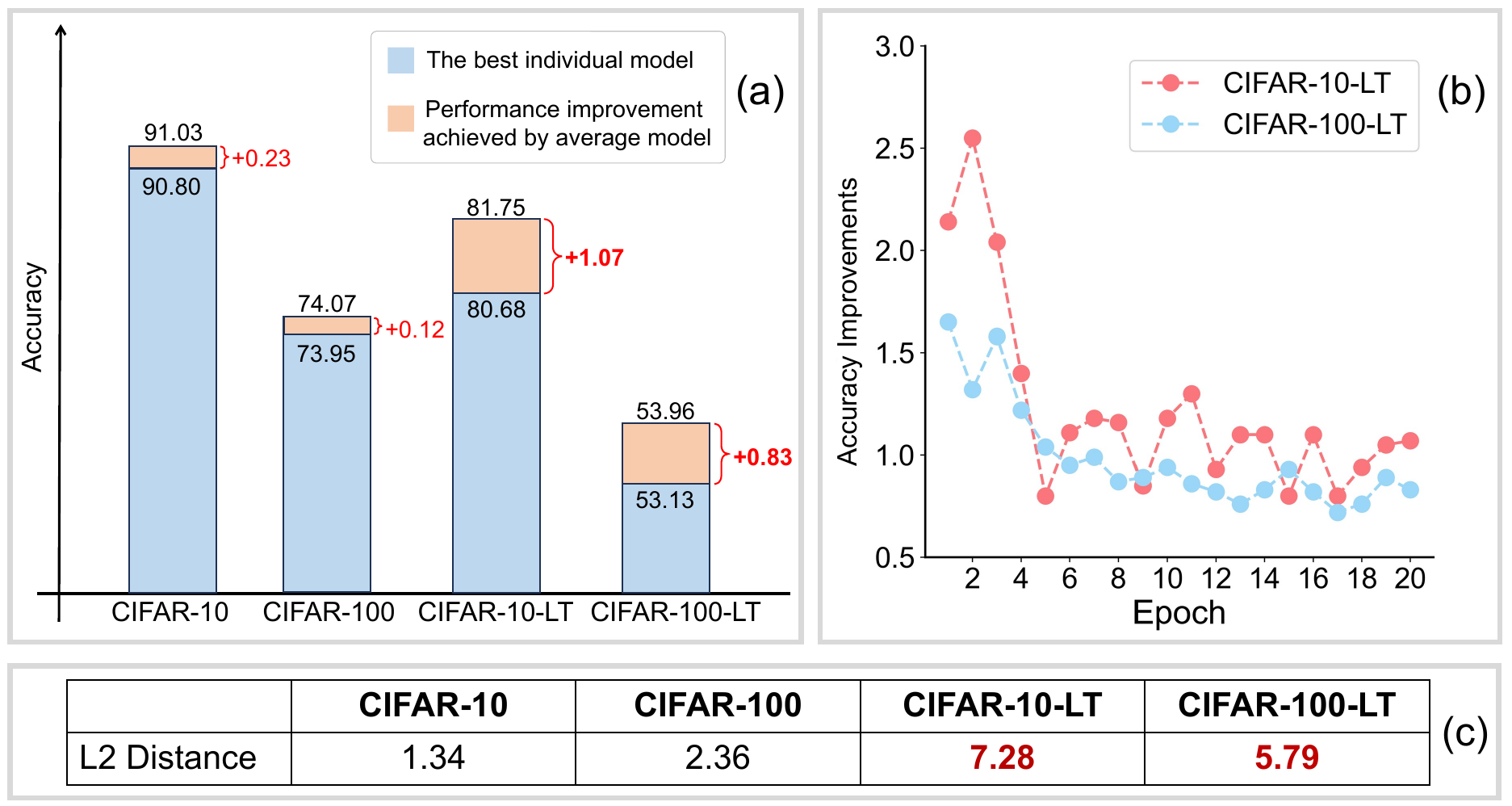}
    \vspace{-1.9em}
    \end{center}

    \caption{\textbf{Observations} from applying MWA on image classification task. 
    The averaged model is obtained by two trained individual ResNet-34~\cite{He_2016_CVPR}.
    \textbf{(a)} \textbf{Accuracy} of the best individual model \textit{vs}. average model on both class-balanced (\ie, CIFAR-10 and CIFAR-100) and class-imbalanced (\ie, CIFAR-10-LT and CIFAR-100-LT) datasets, where average model performs better on the class-imbalanced scenario.  \textbf{(b)} \textbf{Improvements} of the {{average model}} versus each epoch, where performing model weight averaging in the early stage of training brings higher improvement. \textbf{(c)} \textbf{L2 distance} between two trained individual models is higher in the class-imbalanced datasets.
    } 

    \label{fig:fig3}
    \end{figure}

In this work, we empirically deriving two observations from applying MWA on image classification task.
\textbf{1)} The first observation is that \textit{the degree of class balance in the training set affects the performance improvement by MWA}.
As shown in Fig.~\ref{fig:fig3}(a), the average model performs better than the best individual model across all selected datasets, while it shows more significant performance improvements on CIFAR-10-LT and CIFAR-100-LT, both of which are class-imbalanced datasets.
We infer that two models trained on class-imbalanced datasets exhibit greater model's diversity (See Fig.~\ref{fig:fig3}(c)) than that on class-balanced datasets, hence the performance of their average model shows a more pronounced improvement~\cite{liu2022cycle}.
\textbf{2)} The second observation is that \textit{performing model averaging in the early epochs of training yields a greater performance improvement than doing that in later epochs}.
Experimentally, we perform model weight averaging at different epochs respectively, and the results are shown in  Fig.~\ref{fig:fig3}(b).
For example, on CIFAR-10-LT (the red line), 
It achieved over a $\textbf{2\%}$ performance improvement when performing the weight average between epoch 1 and epoch 3, but achieved only about a 1\% improvement near the end of training. 
This phenomenon can also be observed from the results on CIFAR-100-LT.

%
%

From these two observations, one can notice that MWA may be more beneficial for class-imbalanced learning tasks. 
Moreover, performing weight averaging early during training can maximizes the extent of performance improvement. 
These observations have inspired us that, if we use the early averaged model as a new initialization and continue training, iteratively repeating the ``training-average" process, could it yield better results than just performing weight averaging once?
To this end, in this paper we propose a novel MWA technique for class-imbalanced learning named \textbf{I}terative \textbf{M}odel \textbf{W}eight \textbf{A}veraging (IMWA), which iterates the parallel training and weights average process for many times.
%
%
%
%
%
Specifically, IMWA splits the whole training stage into several episodes.
In each episode, multiple models are trained in parallel from the same initialization model weight, with same training iterations but different data sampling orders, and then they are average into one model.
Next, the weight of the average model is treated as a new initialization of the next episodes, which formulates an ``iterative'' manner.
Compared to the vanilla MWA approaches, IMWA enables model to obtain performance improvement from weight average operation at each episode, while incurs almost no additional computational overhead due to the low computational complexity of the average operation. 

We select three distinct class-imbalanced vision tasks to assess effectiveness of IMWA: class-imbalanced image classification, class-imbalanced semi-supervised image classification, and semi-supervised object detection.
In the context of class-imbalanced semi-supervised classification, the labeled dataset exhibits class imbalance.
Furthermore, the model dynamically generates pseudo labels for unlabeled images, resulting in an often uncontrollable number of pseudo labels for each class.
while for the setting of semi-supervised object detection tasks, the common object detection datasets (e.g. \textbf{MS-COCO}) are naturally class-imbalanced.
%
In each task, we adopt IMWA on several state-of-the-art (SoTA) methods of corresponding tasks and conduct experiments on various benchmarks for evaluation. 
%
%
Furthermore, considering that the latter two tasks often employ the Exponential Moving Average (EMA) technique~\cite{liu2021unbiased,oh2022daso} to stabilize training and obtain better performance, we also explore the synergistic collaboration of IMWA and EMA. 
Extensive experimental results show that our IMWA is more beneficial than vanilla MWA in improving performance for these methods, meanwhile, the performance improvement achieved by combining IMWA and EMA surpasses that of using IMWA or EMA individually.
In conclusion, the contributions of this paper are summarized as follows:
\begin{itemize}

    \item [1)] We empirically observe that vanilla MWA performs well in class-imbalanced learning, and performing model averaging in the early epochs of training yields a greater performance improvement than doing that in later epochs, which inspire us to propose a novel MWA technique for class-imbalanced learning tasks named IMWA.
 %
   \item [2)] IMWA executes the multiple model parallel training and weights averaging process in an iterative manner. In addition, we further adopt IMWA to methods that utilize EMA techniques, thus demonstrating that IMWA and EMA complement each other.
   \item [3)] Extensive experiments on various benchmarks of different class-imbalanced learning tasks demonstrate that the IMWA method is superior to the vanilla MWA and can effectively improve the performance of models. 
\end{itemize}

\section{Related Works}
\label{Related_Work}
\subsection{Model Weight Averaging} 
Model weight averaging aims at averaging weights of multiple trained models to improve performance.
The existing model averaging approaches can be concluded into two branches. 
The first kind of methods \cite{szegedy2016rethinking,frankle2020linear,wortsman2022model,ilharco2022patching,rame2022diverse,matena2022merging} trained multiple models simultaneously, and then they average the weights of each converged model to a stronger model.
 And the counterparts \cite{izmailov2018averaging, cha2021swad} showed that simple averaging of multiple checkpoints along one training trajectory of SGD with a typical learning rate schedule, can also improve generalization than conventional training. 
Then, several following methods~\cite{sohn2020fixmatch,oh2022daso,liu2021unbiased} involved the Exponential Moving Average (EMA)  approach to enhance performance and training stability.
Averaging models along the dimension of a individual model's training trajectory can effectively reduce the computational cost during training, but insufficient diversity inevitably lead to performance bottlenecks \cite{liu2022cycle}. 
In this paper, our proposed IMWA extend the first branch approaches while incorporating the concept of the second branch that ``average in the training loop'' by iterating the training-averaging process.
In addition, IMWA can further enhance those methods based on EMA, indicating that IMWA and the second branch approaches are complementary. 
\textcolor{black}{Note that some concurrent works~\cite{jain2023dart,zhang2023lookaround} also proposed to  employ model weight averaging during the training process, While we further analyses the performance of IMWA in class-imbalanced learning tasks and explore the the synergistic collaboration of IMWA and EMA technique.}

\subsection{Class-Imbalanced Learning Tasks} 
Class-imbalanced learning tasks can be summarized into:
\begin{itemize}

    \item [1)] Class-Imbalanced Image Classification. This task aims to train an image classifier on a class-imbalanced training dataset.
Vanilla training with cross-entropy loss may result in overfiting issue on head classes.
To address the issue,  \cite{cui2019class,li2022long,yu2022re,zhou2020bbn} adjusted sampling rate for each class to re-balanced the training data for model. 
\cite{alshammari2022long,park2021influence,tan2020equalization,tang2020long,zhong2021improving} mitigated the class-imbalanced problem  by assigning different class weights for training data.
\cite{xiang2020learning,zhang2021test,menon2020long,li2022nested} applied single expert learning and knowledge aggregation to alleviate the class-imbalanced problem.
\cite{ma2021simple,tian2022vl,long2022retrieval} proposed to involve the vision-language model, introducing  language knowledge and external database to adjust the output scores.
Recently, \cite{xu2023learning,dong2022lpt} involved ViT to alleviate the class-imbalanced problem by propose masked generative pretraining or prompts tunning. Different from these methods above, IMWA aims to enhance the model's generalization ability from the perspective of averaging model parameters.
Therefore, IWMA is plug-and-play approach and applied to these existing methods easily without modifying their origin structure or training details.
   \item [2)] Semi-Supervised Class-Imbalanced Image Classification. This task aims to train an image classifier on a labeled set and unlabeled set, both of which are suffered from the class-imbalanced problem~\cite{wei2023towards,oh2022daso}. In addition, pseudo labels on the unlabeled set are changeable
 during training, which can also lead to class imbalance.
To tackle this problem, \cite{wei2021crest,kim2020distribution} assumed that the class distribution in the labeled and unlabeled sets are consistent, and using this assumption to control the generation of pseudo-labels for unlabeled data.
\cite{lee2021abc} added an extra balanced classifier, which is trained to be balanced across all classes by using a mask that re-balances the class distribution.
While \cite{oh2022daso,wei2023towards} proposed that the class distribution of labeled set and unlabeled set are different, and present dynamic or adaptive manner to learn the truth distribution. 
We applied IMWA to several SOTA approaches of this task, which demonstrated that IMWA is also beneficial for semi-supervised learning scenarios in terms of performance.
Notice that most approaches use EMA (Exponential Moving Average) technique to maintain a model with better generalization.
Therefore, in this paper we also propose a improved  version of IMWA for collaborating with EMA, which demonstrating that IMWA and EMA complement each other.
   \item [3)] Semi-Supervised Object Detection. This task aims to train an object detector on a full-annotated labeled set combining with a huge unlabeled set.
Due to the imbalance in the number of instances for different classed in most common object detection datasets, this task also suffers from class-imbalanced problem.
Recent mainstream methods presented the teacher-student training framework~\cite{liu2021unbiased,xu2021end,chen2022dense,zhou2022dense,liu2022unbiased,li2022dtg,chen2022label,kim2022mum,liu2022cycle,li2022semi,zhang2023controlvideo}, where the student is optimized with pseudo-labels generated by teacher model while the weights of teacher model are updated by student model gradually with EMA.
In this paper, we also apply our IMWA on existing SSOD approaches to IMWA is benefit not only for image classification task, but also for object detection field.

\end{itemize}

\begin{figure*}[!htb]

    \begin{center}
    \includegraphics[width=1.0\linewidth]{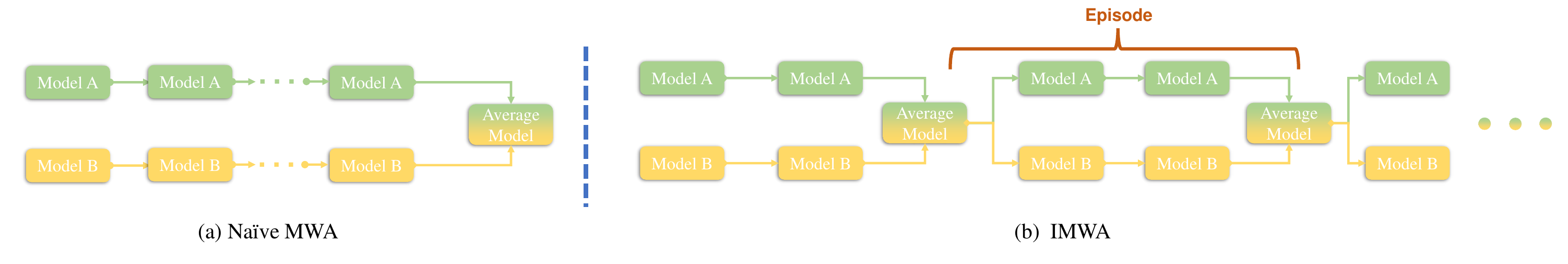}
    \end{center}
    \caption{The \textbf{illustration} of \textbf{(a)} vanilla Model Weight Averaging,  and \textbf{(b)} our proposed \textbf{Iterative Model Weight Averaging (IMWA)}.  IMWA splits the whole training stage into several episodes. At each episode, multiple models are trained in parallel from a same initialization model weight, with same training iterations but different data orders, and then they are average into one model.
Then, the weight of the average model will be treated as a new initialization for the next episodes, which forms a ``iterative'' manner.} 
    \label{fig:pipeline}
    \end{figure*}

\section{Method}\label{sec:method}
\subsection{Vanilla MWA}
The vanilla MWA approach~\cite{wortsman2022model} consists of two steps: the parallel training step and the averaging step.
Given a set of models $\{\bm{f}(\mathbf{x};\bm{\theta}_1,\mathcal{H}_1),...,\bm{f}(\mathbf{x};\bm{\theta}_M,\mathcal{H}_M)\}$, $\mathbf{x}$ denotes the input data, $M$ denotes the number of models involved; $\bm\theta_m$ denotes model weights and the $\mathcal{H}_m$ denotes a group of training hyper-parameters, where $m \in [1,...,M]$.
Note that all these models share the same network architecture, while their weights $\bm \theta_m$ and corresponding training hyper-parameter $\mathcal{H}_m$ may be different. 

During the parallel training step, all models' weights are assigned by the same initialization. 
%
%
Then they are trained in parallel. 
And during the averaging step, corresponding weights trained models are averaged by Eq.~(\ref{eq:mwa}),
\begin{equation}\label{eq:mwa}
    \bm{\theta}_{{\rm MWA}} = \sum_{m=1}^{M} \alpha_{m} \bm{\theta}_{m},
\end{equation}
where $\bm{\theta}_{{\rm MWA}}$ denotes the weights of average model, $\bm{\theta}_{m}$ denotes the weights of a individual trained model, $\alpha_m$ denotes the coefficient for $\bm{\theta}_{m}$ which satisfies $\sum_{m=1}^{M} \alpha_{m}=1$, and $M$ denotes the total number of models. 
In the inference stage, only $\bm{\theta}_{{\rm MWA}}$ will be evaluated to obtain the performance score.

Previous MWA methods~\cite{wortsman2022model,matena2022merging,rame2022diverse} typically involve training each model to convergence before averaging their weights.
While our empirical findings suggest that balancing the parameters early in the training of each model (for example, during the first few epochs) results in a more significant performance improvement in the averaged model compared to averaging after convergence, as mentioned in Sec.~\ref{sec:intro}.
This inspires us to consider whether the averaged model at early epoch could serve as a new starting point for training, and then continuously iterate this process of "training" and "weight averaging." 
To this end, we propose Iterative Model Weight Averaging.

\subsection{Iterative Model Weight Averaging}
\subsubsection{Overview} \textcolor{black}{The key idea of IMWA is employing the vanilla MWA approach during the training process, which treats the average model as a new initialization point to repeat the vanilla MWA process for many times.} 
%
We refer to each iteration process as an ``episode'' and split the whole training process into $E$ episodes. 
The pseudo code of IMWA can be refer to Algorithm~\ref{alg4}. And corresponding illustration is shown in Fig~\ref{fig:pipeline}.
Specifically, the process of our IMWA can be divided into the following key components:
%
\subsubsection{Initialization}
At the beginning of the whole training process, the weights of $M$ individual models are all initialized with the same weights $\bm{\theta}^{(0)}$:

\begin{equation}\label{eq:init}
       \bm{\theta}_{m}^{(0)} \gets \bm{\theta}^{(0)},
\end{equation}
where $m \in \{1,...,M\}$. 
Note that $\bm{\theta}^{(0)}$ can be obtained by random initialization (for CIIC and CISSIC tasks) or pretraining backbone (for SSOD task). 
Similar to vanilla MWA, we introduce a unique hyper-parameter set $\mathcal{H}_{m}$ for each model for optimization during training. 

 \begin{algorithm}[!t]
    \SetAlgoLined
	\caption{Iterative Model Weight Averaging}
	\label{alg4}
	{\bfseries Input:}  A specific algorithm $\mathcal{A}(\cdot)$; the initialization weights $\bm{\theta}^{(0)}$; training hyper-parameters $\{\mathcal{H}_{1},...,\mathcal{H}_{M}\}$; 
    the number of episode $E$; the number of training iterations of  in each episode $T_e$; the number of models $M$; the training dataset $\mathcal{D}$\;
    {\bfseries Output:} the average weights $\bm{\theta}^{(E)}$ after training\;
   \For{$e=1,\cdots, E$}{
       \For{$m=1,\cdots, M$}{
     Update  $\bm{\theta}_{m}^{(e-1)} \gets \bm{\theta}^{(e-1)}$\;
    Compute $\bm{\theta}_{m}^{(e-1)'}
    \gets \mathcal{A}(\bm{\theta}_{m}^{(e-1)},\mathcal{D},\mathcal{H}_{m},T_e)$\;
    
    }
Compute $\bm{\theta}^{(e)} = \frac{1}{M}\sum_{m=1}^{M}  \bm{\theta}_{m}^{(e-1)'}$\;

   }
   Return $\bm{\theta}^{(E)}$

\end{algorithm}

\subsubsection{Episode} Similar to the vanilla MWA, each episode consists of two steps: parallel training step and weight average step. 
At the parallel training step, $M$ models are trained in parallel via a specific algorithm $\mathcal{A}(\cdot)$ on a certain dataset $\mathcal{D}$ \textcolor{black}{for $T_e$ iterations, where $T_e=\frac{T}{E}$ and $T$ denotes the total number of training iterations of whole training stage.}

Formally, we leverage Eq.~(\ref{eq:pa}) to optimize each model $\bm f$:

\begin{equation}\label{eq:pa}
    \bm{\theta}_{m}^{(e-1)'}
    \gets \mathcal{A}(\bm{\theta}_{m}^{(e-1)},\mathcal{D},\mathcal{H}_{m},T_e),
\end{equation}
where the superscript $e$ represents the index of episode.
At the weight average step, we average these trained weights $\bm{\theta}_{m}^{(e-1)'}$ by Eq.~(\ref{eq:aver}):
\begin{equation}\label{eq:aver}
    \bm{\theta}^{(e)} = \sum_{m=1}^{M}  \alpha_m \bm{\theta}_{m}^{(e-1)'},
\end{equation}
\textcolor{black}{where the coefficient $\alpha_m$ referred in vanilla MWA is set to $\frac{1}{M}$  to avoid cumbersome manual tuning.}

\subsubsection{Iterative loop}
After obtaining the average weight  $\bm{\theta}^{(e)}$, we re-assign it to each model:
\begin{equation}\label{eq:update}
       \bm{\theta}_{m}^{(e)} \gets \bm{\theta}^{(e)}.
\end{equation}
Now each model has updated with new weights $\bm{\theta}_{m}^{(e)}$, \textcolor{black}{then conduct the next episode training.}
\subsubsection{Evaluation}
Upon whole training completion, we take the average model of the last episode $\bm{\theta}^{(E)}$ as the final model and then evaluate its performance on the test set.

\subsection{Collaboration with EMA}
Now we have introduced the process of IMWA in the previous sub-section, where we assume that only optimized model $\bm f$ is involved to the selected SOTA algorithm $\mathcal{A}(\cdot)$.
However, many recent works~\cite{oh2022daso,wei2023towards,liu2021unbiased,menendez1997jensen} of CISSIC and SSOD adopt the EMA strategy to improved effectiveness and stability.
In these works, an extra EMA model $\bm f(\textbf{x};\bm\omega,\mathcal{H})$ is updated at the $t$ iteration with the optimized model $\bm f$ by Eq.~(\ref{eq:ema})
\begin{equation}\label{eq:ema}
       \bm{\omega}^t = \lambda\bm{\omega}^{t-1}+(1-\lambda)\bm{\theta}^t,
\end{equation}
where $\lambda$ is the coefficient and usually set to close to 1. 
Involving the EMA model necessitates modifications to our IMWA.
To extend our IMWA to these adopting EMA methods, we propose the following modifications: 
\begin{itemize}

    \item [1)] To merge the EMA strategy into the baseline method $\mathcal{A}(\cdot)$, we modify the Eq.~(\ref{eq:pa}) to Eq.~(\ref{eq:pa_ema}):
 \begin{equation}\label{eq:pa_ema}
    \bm{\theta}_{m}^{(e-1)'},\bm{\omega}_{m}^{(e-1)'}
    \gets \mathcal{A}(\bm{\theta}_{m}^{(e-1)},\bm{\omega}_{m}^{(e-1)},\mathcal{D},\mathcal{H}_m,T_e).
\end{equation}   
    
    \item [2)] Then we extend Eq.~(\ref{eq:aver}) and Eq.~(\ref{eq:update}) for the weights of EMA models:
\begin{equation}\label{eq:aver_ema}
    \bm{\omega}^{(e)} = \frac{1}{M}\sum_{m=1}^{M}  \bm{\omega}_{m}^{(e-1)'},
\end{equation}

\begin{equation}\label{eq:1_ema}
       \bm{\omega}_{m}^{(e)} \gets \bm{\omega}^{(e)}.
\end{equation}
   \item [3)] Upon whole training completion, we take the average EMA model of the last episode $\bm{\omega}^{(E)}$ as the final model and then evaluate its performance on the test set.

\end{itemize}

\begin{table}[t]
    \centering
    \caption{\textbf{Performance} of our IMWA applying to BCL~\cite{zhu2022balanced} and LiVT~\cite{xu2023learning} on \textbf{three CIIC  benchmarks}. ``\textbf{IN-LT}'' is the abbreviation for ``\textbf{ImageNet-LT}'' and ``\textbf{iNat-18}'' is the abbreviation for iNaturalist 2018. }
 
    \label{tab:compare_w_sota_ciic}
    \centering
    \fontsize{8.3}{11}\selectfont
    \setlength{\tabcolsep}{15pt}
    \begin{tabular}{l| c  c  c}

    \toprule
    \textbf{Methods}
    &\textbf{IN-LT} 
    &\textbf{iNat18} 
    &\textbf{Places-LT}\\
    \cmidrule(lr){1-1}  \cmidrule(lr){2-2} \cmidrule(lr){3-3} \cmidrule(lr){4-4} 
    ACE~\cite{cai2021ace}
    &$56.6$ 
    &$72.9$ 
    &- \\
    
    PaCo~\cite{cui2021parametric}
    &$58.2$ 
    &$73.2$ 
    &$41.2$ \\
    
    TADE~\cite{zhang2021test}
    &$\textbf{58.8}$ 
    &$72.9$
    &$40.9$ \\

    TSC~\cite{li2022targeted}
    &$52.4$ 
    &$69.7$
    &- \\
    
    GCL~\cite{li2022long}
    &$54.5$
    &$71.0$
    &$40.6$ \\
    
     TLC~\cite{li2022trustworthy}
     &$55.1$
     &- 
     &-\\
     
     NCL~\cite{li2022nested}
     &$57.7$
     &$\textbf{74.2}$
     &$\textbf{41.5}$ \\
     
     Bread~\cite{liu2022breadcrumbs}
     &$44.0$
     &$70.3$
     &$39.3$ \\
     
     DOC~\cite{wang2022towards}
     &$55.0$
     &$71.0$
     &- \\

     DLSA~\cite{xu2022constructing}
     &$57.5$
     &$72.8$
     &$39.0$ \\
 
    \cmidrule(lr){1-1}  \cmidrule(lr){2-2} \cmidrule(lr){3-3} \cmidrule(lr){4-4} 
    
    \cellcolor{mypink2}BCL~\cite{zhu2022balanced}
    &\cellcolor{mypink2}$56.0$
    &\cellcolor{mypink2}$71.8$ 
    &\cellcolor{mypink2}$39.4$ \\
 
    \cellcolor{mypink}\textbf{BCL+IMWA}
    &\cellcolor{mypink}$\textbf{57.3}$
    &\cellcolor{mypink}$\textbf{72.5}$
    &\cellcolor{mypink}$\textbf{40.3}$ \\

    \cellcolor{mypink2}LiVT~\cite{xu2023learning}
    &\cellcolor{mypink2}$58.2$
    &\cellcolor{mypink2}$75.8$
    &\cellcolor{mypink2}$35.1$\\  

    \cellcolor{mypink}\textbf{LiVT+IMWA}
    &\cellcolor{mypink}$\textbf{59.0}$
    &\cellcolor{mypink}$\textbf{76.6}$
    &\cellcolor{mypink}$\textbf{36.7}$\\

    \bottomrule
    \end{tabular}

\end{table}

\begin{figure}[htb]
    \begin{center}
    \includegraphics[width=1.0\linewidth]{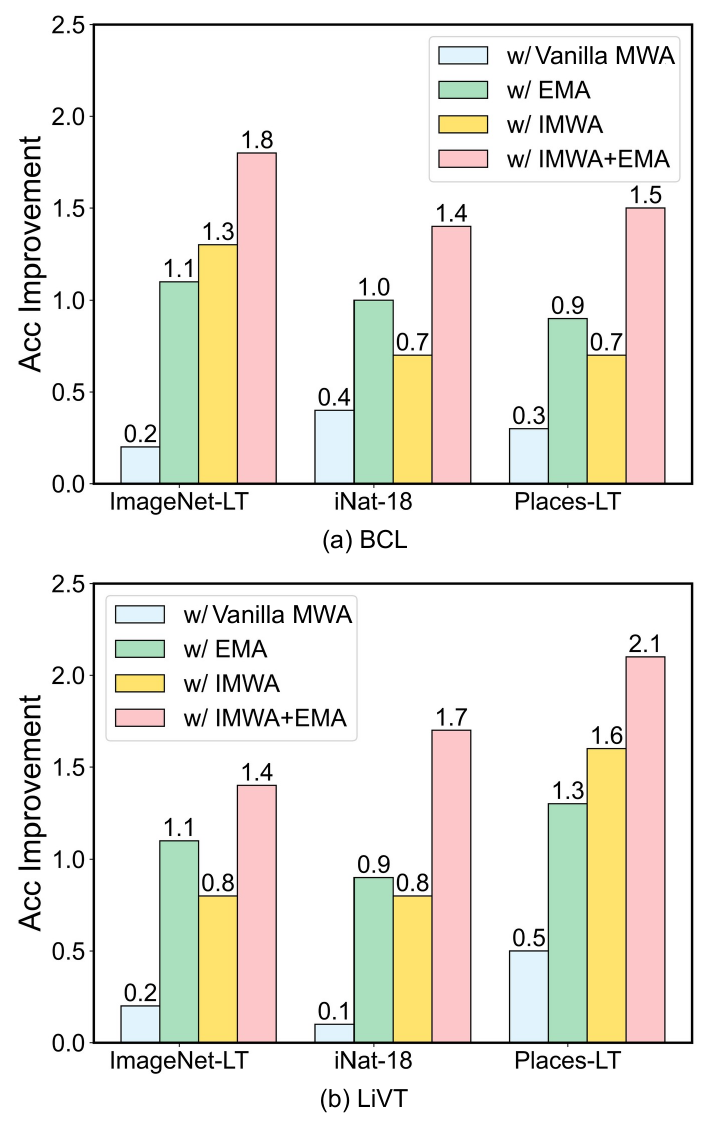}
    \end{center}
	\caption{\textbf{Comparison} of our IMWA and other MWA approaches in terms of achieving performance improvements for \textbf{(a)} BCL~\cite{zhu2022balanced} and \textbf{(b)} LiVT~\cite{xu2023learning} on \textbf{three CIIC benchmarks}.}

    \label{fig:ciic_ema}
\end{figure}

\section{Experiments}\label{sec:exp}

\subsection{Evaluation Tasks and Experimental Setup}
This paper selects three distinct class-imbalanced vision tasks to assess effectiveness of IMWA: class-imbalanced image classification (CIIC), class-imbalanced semi-supervised image classification (CISSIC), and semi-supervised object detection (SSOD).
The problem definitions of these tasks are presented below:
\begin{itemize}
	\item CIIC aims to training an image classifier with a class-imbalanced dataset $\mathcal{D}_{\rm CI}$. 
 In formal, $\mathcal{D}_{\rm CI}=\{(\mathbf{x}_i,y_i)\}^{N_{\rm CI}}_{i=1}$. $\mathbf{x}_i \in \mathbb{R}^{w \times h \times 3}$ denotes an RGB image, where $w$ and $h$ indicate the width and height of image.
 $y_i \in \{1,...,C\}$ denotes the label of $\mathbf{x}_i$ where $C$ is the total number of class in $\mathcal{D}_{\rm CI}$. 
 $N_{\rm CI}=\sum^{C}_{c=1} n_c$ denotes the total number of training pairs, where $n_c$ denotes the of number training pairs belonging to the class $c$. 
 In this paper we assume the $\{n_c\}^C_{c=1}$ obeys a long-tailed distribution, which is commonly adopted in previous works. 
 We set $n_1 \geq n_2 \geq \cdots \geq n_C$ and imbalance ratio $\gamma = \frac{n_1}{n_C}$ to represent the degree of imbalance.

	\item CISSIC aims to training an image classifier with a labeled dataset $\mathcal{D}_{\rm CISS}^l=\{(\mathbf{x}_i,y_i)\}^{N_{\rm CISS}^l}_{i=1}$ and an unlabeled dataset $\mathcal{D}_{\rm CISS}^u=\{\mathbf{x}_i\}^{N^u_{\rm CISS}}_{i=1}$, where $N_{\rm CISS}^l$ and $N^u_{\rm CISS}$ denotes the number of images in $\mathcal{D}_{\rm CISS}^l$ and $\mathcal{D}_{\rm CISS}^u$ respectively. 
 Similar to CIIC, both $\mathcal{D}_{\rm CISS}^l$ and  $\mathcal{D}_{\rm CISS}^u$ are class-imbalanced.
 we set $\gamma^l$ and $\gamma^u$ as the imbalance ratios for these two sets.
	\item SSOD aims to training an object detection model in a semi-supervised manner, in which a set of full-labeled images $\mathcal{D}^l_{\rm SSOD}=\{(\mathbf{x}_i,\mathcal{B}_i)\}_{i=1}^{N^{l}_{\rm SSOD}}$ combination with a set of unlabeled images $\mathcal{D}^u_{\rm SSOD}=\{\mathbf{x}_j\}_{j=1}^{N^{u}_{\rm SSOD}}$ are available.   The $\mathcal{B}_i = \{(\mathbf{b}_k,y_k)\}_{k=1}^{K_i}$ denotes the box-level annotations of $\mathbf{x}_i$ consisting of $K_i$ bounding box labels, where $\mathbf{b}_k$ is the $k$-th bounding box coordinates, and $y_k \in \{1,...,C\}$ is the class label of $\mathbf{b}_k$. 
\end{itemize}

For each evaluation task, we apply IMWA to several recent SOTA methods , and then select various widely used benchmarks to evaluate the performance.
Specifically:
\begin{itemize}
	\item For CIIC, we apply IMWA to LiVT \cite{xu2023learning} and BCL \cite{zhu2022balanced}.
 We evaluate our IMWA for CIIC on ImageNet-LT \cite{liu2019large,deng2009imagenet}, iNaturalist 2018 (iNat18) \cite{van2018inaturalist} and Places-LT \cite{zhou2017places} benchmarks.
 
	\item For CISSIC, we apply IMWA to ACR \cite{wei2023towards} and DASO \cite{oh2022daso}.
 We evaluate our IMWA for CISSIC on CIFAR-10/100-LT \cite{krizhevsky2009learning}, STL-10\cite{coates2011analysis} and ImageNet-127 \cite{huh2016makes} benchmarks.
	\item For SSOD, we apply IMWA to Dense Teacher \cite{zhou2022dense} , Soft Teacher \cite{xu2021end} and Unbiased Teacher \cite{liu2021unbiased}.
  We evaluate our IMWA for SSOD on MS-COCO \cite{lin2014microsoft}, PASCAL VOC \cite{everingham2010pascal}  benchmarks.
\end{itemize}

\subsection{Implementation Details}
In this subsection we introduce some implementation details of our experiments.
As mentioned above, we apply our IMWA to several SOTA methods for each task. 
For each method, we deploy our IMWA on its official implementation and follow its default training configurations (\eg  total training iteration, batch size, data augmentation, and other hyper-parameters) for fair comparison.
For the hyper-parameters of our IMWA, we set $E$ to 20 and $M$ to 2 for all experiments in default.
 For the configurations of training components $\{\mathcal{H}_{1},...,\mathcal{H}_{M}\}$, we simply maintain an independent training dataloader for each individual model. 
 Therefore, each model is trained with different sequence of sample data and optimizing direction during training period. 
 For CIIC task and CISSIC task, $\bm{\theta}^{(0)}$ is initialized from the scratch.
 While for SSOD task, the backbone (feature extractor) of $\bm{\theta}^{(0)}$ is pretrained from ImageNet.
 The experiments of applying IMWA on Soft Teacher~\cite{xu2021end} are deployed on 8 $\times$ Tesla V100, and the other experiments are deployed on  8 $\times$ RTX 2080 Ti.

Besides comparing with these SOTA methods, we compare our IMWA with these performing the vanilla MWA and EMA approaches.
For vanilla MWA approach,  we perform it  for all three tasks, just like the experiments of our IMWA.
While as for EMA,  because the methods for CISSIC and SSOD inherently incorporate EMA techniques (Please refer to their works for details), thus our experiments for these two tasks mainly explore the effect of synergistic collaboration between IMWA and EMA.
To further present the comparison of IMWA and EMA, we conducted extra experiments for EMA on CIIC task.
In addition, for each task, we also compare our IMWA with other recent influential methods to further show the effectiveness of IMWA.

\begin{table*}[htb]
\centering
\caption{
\textbf{Performance} of our IMWA applying to DASO~\cite{oh2022daso} and ACR\cite{wei2023towards} under consistent class distributions on \textbf{CISSIC benchmarks}, \ie, \textbf{CIFAR10-LT} and \textbf{CIFAR100-LT}.
}
     \fontsize{8.3}{11}\selectfont
    \setlength{\tabcolsep}{7pt}
    \begin{tabular}{lcccccccc}
         \toprule
         &\multicolumn{4}{c}{\textbf{CIFAR10-LT}} 
         &\multicolumn{4}{c}{\textbf{CIFAR100-LT}} \\
         
         &\multicolumn{2}{c}{$\gamma=\gamma_l=\gamma_u=100$}
         &\multicolumn{2}{c}{$\gamma=\gamma_l=\gamma_u=150$}
         &\multicolumn{2}{c}{$\gamma=\gamma_l=\gamma_u=10$}
         &\multicolumn{2}{c}{$\gamma=\gamma_l=\gamma_u=20$} \\
         
         \cmidrule(lr){2-3} \cmidrule(lr){4-5} \cmidrule(lr){6-7} \cmidrule(l){8-9}
         
         \multirow{2}{*}{\textbf{Algorithm}}
         &$N_1=500$
         &$N_1=1500$
         &$N_1=500$
         &$N_1=1500$
         &$N_1=50$
         &$N_1=150$
         &$N_1=50$
         &$N_1=150$ \\
         
         &$M_1=4000$
         &$M_1=3000$
         &$M_1=4000$
         &$M_1=3000$
         &$M_1=400$
         &$M_1=300$
         &$M_1=400$
         &$M_1=300$ \\

        \cmidrule(r){1-1} \cmidrule(lr){2-3} \cmidrule(lr){4-5} \cmidrule(lr){6-7} \cmidrule(l){8-9}

        Supervised 
        &\ms{$47.3$}{$0.95$}
        &\ms{$61.9$}{$0.41$}
        &\ms{$44.2$}{$0.33$}
        &\ms{$58.2$}{$0.29$}
        &\ms{$29.6$}{$0.57$}
        &\ms{$46.9$}{$0.22$}
        &\ms{$25.1$}{$1.14$}
        &\ms{$41.2$}{$0.15$} \\
        
        ~~ w/ LA~\cite{menon2020long}
        &\ms{$53.3$}{$0.44$}
        &\ms{$70.6$}{$0.21$}
        &\ms{$49.5$}{$0.40$}
        &\ms{$67.1$}{$0.78$}
        &\ms{$30.2$}{$0.44$}
        &\ms{$48.7$}{$0.89$}
        &\ms{$26.5$}{$1.31$}
        &\ms{$44.1$}{$0.42$} \\

        \cmidrule(r){1-1} \cmidrule(lr){2-3} \cmidrule(lr){4-5} \cmidrule(lr){6-7} \cmidrule(l){8-9}

        FixMatch + LA~\cite{menon2020long}
        &\ms{$75.3$}{$2.45$}
        &\ms{$82.0$}{$0.36$}
        &\ms{$67.0$}{$2.49$}
        &\ms{$78.0$}{$0.91$}
        &\ms{$47.3$}{$0.42$}
        &\ms{$58.6$}{$0.36$}
        &\ms{$41.4$}{$0.93$}
        &\ms{$53.4$}{$0.32$} \\
        
        ~~w/ DARP~\cite{kim2020distribution}
        &\ms{$76.6$}{$0.92$}
        &\ms{$80.8$}{$0.62$}
        &\ms{$68.2$}{$0.94$}
        &\ms{$76.7$}{$1.13$}
        &\ms{$50.5$}{$0.78$}
        &\ms{$59.9$}{$0.32$}
        &\ms{$44.4$}{$0.65$}
        &\ms{$53.8$}{$0.43$} \\
        
        ~~w/ CReST+~\cite{wei2021crest}
        &\ms{$76.7$}{$1.13$}
        &\ms{$81.1$}{$0.57$}
        &\ms{$70.9$}{$1.18$}
        &\ms{$77.9$}{$0.71$}
        &\ms{$44.0$}{$0.21$}
        &\ms{$57.1$}{$0.55$}
        &\ms{$40.6$}{$0.55$}
        &\ms{$52.3$}{$0.20$} \\
        
        ~~w/ DASO~\cite{oh2022daso}
        &\ms{$77.9$}{$0.88$}
        &\ms{$82.5$}{$0.08$}
        &\ms{$70.1$}{$1.68$}
        &\ms{$79.0$}{$2.23$}
        &\ms{$50.7$}{$0.51$}
        &\ms{$60.6$}{$0.71$}
        &\ms{$44.1$}{$0.61$}
        &\ms{$55.1$}{$0.72$} \\
        
        \cmidrule(r){1-1} \cmidrule(lr){2-3} \cmidrule(lr){4-5} \cmidrule(lr){6-7} \cmidrule(l){8-9}

        FixMatch + ABC~\cite{lee2021abc}
        &\ms{$78.9$}{$0.82$}
        &\ms{$83.8$}{$0.36$}
        &\ms{$66.5$}{$0.78$}
        &\ms{$80.1$}{$0.45$}
        &\ms{$47.5$}{$0.18$}
        &\ms{$59.1$}{$0.21$}
        &\ms{$41.6$}{$0.83$}
        &\ms{5$3.7$}{$0.55$} \\
        
        ~~w/ DASO~\cite{oh2022daso} 
        &\ms{$80.1$}{$1.16$}
        &\ms{$83.4$}{$0.31$}
        &\ms{$70.6$}{$0.80$}
        &\ms{$80.4$}{$0.56$}
        &\ms{$50.2$}{$0.62$}
        &\ms{$60.0$}{$0.32$}
        &\ms{$44.5$}{$0.25$}
        &\ms{$55.3$}{$0.53$} \\
        
 \cmidrule(r){1-1} \cmidrule(lr){2-3} \cmidrule(lr){4-5} \cmidrule(lr){6-7} \cmidrule(l){8-9}
 
        FixMatch~\cite{menon2020long} 
        &\ms{$67.8$}{$1.13$}
        &\ms{$77.5$}{$1.32$}
        &\ms{$62.9$}{$0.36$}
        &\ms{$72.4$}{$1.03$}
        &\ms{$45.2$}{$0.55$}
        &\ms{$56.5$}{$0.06$}
        &\ms{$40.0$}{$0.96$}
        &\ms{$50.7$}{$0.25$}\\
        
        \cellcolor{mypink2}~~w/ DASO ~\cite{oh2022daso}
        &\cellcolor{mypink2}\ms{$76.0$}{$0.37$}
        &\cellcolor{mypink2}\ms{$79.1$}{$0.75$}
        &\cellcolor{mypink2}\ms{$70.1$}{$1.81$}
        &\cellcolor{mypink2}\ms{$75.1$}{$0.77$}
        &\cellcolor{mypink2}\ms{$49.8$}{$0.24$}
        &\cellcolor{mypink2}\ms{$59.2$}{$0.35$}
        &\cellcolor{mypink2}\ms{$43.6$}{$0.09$}
        &\cellcolor{mypink2}\ms{$52.9$}{$0.42$} \\

        \cellcolor{mypink}\textbf{~~w/ DASO+IMWA}
        &\cellcolor{mypink}\ms{$\textbf{76.9}$}{$\textbf{0.14}$}
        &\cellcolor{mypink}\ms{$\textbf{79.5}$}{$\textbf{0.37}$}
        &\cellcolor{mypink}\ms{$\textbf{71.3}$}{$\textbf{0.88}$}
        &\cellcolor{mypink}\ms{$\textbf{75.9}$}{$\textbf{0.51}$}
        &\cellcolor{mypink}\ms{$\textbf{50.5}$}{$\textbf{0.36}$}
        &\cellcolor{mypink}\ms{$\textbf{60.1}$}{$\textbf{0.32}$}
        &\cellcolor{mypink}\ms{$\textbf{44.0}$}{$\textbf{0.07}$}
        &\cellcolor{mypink}\ms{$\textbf{53.9}$}{$\textbf{0.27}$} \\
        
        \cellcolor{mypink2}~~w/ ACR ~\cite{wei2023towards}
        &\cellcolor{mypink2}\ms{$81.6$}{$0.19$}
        &\cellcolor{mypink2}\ms{$84.3$}{$0.39$}
        &\cellcolor{mypink2}\ms{$77.0$}{$1.19$}
        &\cellcolor{mypink2}\ms{$80.9$}{$0.22$}
        &\cellcolor{mypink2}\ms{$55.7$}{$0.12$}
        &\cellcolor{mypink2}\ms{$65.6$}{$0.16$}
        &\cellcolor{mypink2}\ms{$48.0$}{$0.75$}
        &\cellcolor{mypink2}\ms{$58.9$}{$0.36$} \\
        
        \cellcolor{mypink}\textbf{~~w/ ACR+IMWA}
        &\cellcolor{mypink}\ms{$\textbf{82.0}$}{$\textbf{0.27}$}
        &\cellcolor{mypink}\ms{$\textbf{84.9}$}{$\textbf{0.55}$}
        &\cellcolor{mypink}\ms{$\textbf{77.7}$}{$\textbf{1.26}$}
        &\cellcolor{mypink}\ms{$\textbf{81.3}$}{$\textbf{0.12}$}
        &\cellcolor{mypink}\ms{$\textbf{56.4}$}{$\textbf{0.45}$}
        &\cellcolor{mypink}\ms{$\textbf{66.2}$}{$\textbf{0.05}$}
        &\cellcolor{mypink}\ms{$\textbf{48.5}$}{$\textbf{0.44}$}
        &\cellcolor{mypink}\ms{$\textbf{59.3}$}{$\textbf{0.19}$} \\
        \bottomrule
    \end{tabular}

\label{tab:CISSIC_1}
\end{table*}

\begin{table*}[htb]
\centering
\caption{\textbf{Performance} of our IMWA applying to DASO~\cite{oh2022daso} and ACR\cite{wei2023towards} under inconsistent class distributions on \textbf{CISSIC benchmarks}, \ie, \textbf{CIFAR10-LT} and \textbf{STL-LT}.
}
\fontsize{8.3}{11}\selectfont
\setlength{\tabcolsep}{7pt}
    \begin{tabular}{lcccccccccc}
        \toprule
         &\multicolumn{4}{c}{\textbf{CIFAR10-LT} ($\gamma_l \neq \gamma_u$)}
         &\multicolumn{4}{c}{\textbf{STL10-LT} ($\gamma_u=$\emph{ N/A})} \\
         
         &\multicolumn{2}{c}{$\gamma_u=1$ (uniform)}
         &\multicolumn{2}{c}{$\gamma_u=1/100$ (reversed)}
         &\multicolumn{2}{c}{$\gamma_l=10$}
         &\multicolumn{2}{c}{$\gamma_l=20$} \\
         
         \cmidrule(lr){2-3} \cmidrule(lr){4-5} \cmidrule(lr){6-7} \cmidrule(l){8-9}
         
         \multirow{2}{*}{\textbf{Algorithm}}
         &$N_1=500$
         &$N_1=1500$
         &$N_1=500$
         &$N_1=1500$
         &$N_1=150$
         &$N_1=450$
         &$N_1=150$
         &$N_1=450$ \\
         
         &$M_1=4000$
         &$M_1=3000$
         &$M_1=4000$
         &$M_1=3000$
         &$M=100k$
         &$M=100k$
         &$M=100k$
         &$M=100k$ \\
         
        \cmidrule(r){1-1} \cmidrule(lr){2-3} \cmidrule(lr){4-5} \cmidrule(lr){6-7} \cmidrule(l){8-9}
        
        FixMatch~\cite{sohn2020fixmatch}
        &\ms{$73.0$}{$3.81$}
        &\ms{$81.5$}{$1.15$}
        &\ms{$62.5$}{$0.94$}
        &\ms{$71.8$}{$1.70$}
        &\ms{$56.1$}{$2.32$}
        &\ms{$72.4$}{$0.71$}
        &\ms{$47.6$}{$4.87$}
        &\ms{$64.0$}{$2.27$} \\
        
        ~~w/ DARP~\cite{kim2020distribution}
        &\ms{$82.5$}{$0.75$}
        &\ms{$84.6$}{$0.34$}
        &\ms{$70.1$}{$0.22$}
        &\ms{$80.0$}{$0.93$}
        &\ms{$66.9$}{$1.66$}
        &\ms{$75.6$}{$0.45$}
        &\ms{$59.9$}{$2.17$}
        &\ms{$72.3$}{$0.60$}\\
        
        ~~w/ CReST~\cite{wei2021crest}
        &\ms{$83.2$}{$1.67$}
        &\ms{$87.1$}{$0.28$}
        &\ms{$70.7$}{$2.02$}
        &\ms{$80.8$}{$0.39$}
        &\ms{$61.7$}{$2.51$}
        &\ms{$71.6$}{$1.17$}
        &\ms{$57.1$}{$3.67$}
        &\ms{$68.6$}{$0.88$} \\
        
        ~~w/ CReST+~\cite{wei2021crest}
        &\ms{$82.2$}{$1.53$}
        &\ms{$86.4$}{$0.42$}
        &\ms{$62.9$}{$1.39$}
        &\ms{$72.9$}{$2.00$}
        &\ms{$61.2$}{$1.27$}
        &\ms{$71.5$}{$0.96$}
        &\ms{$56.0$}{$3.19$}
        &\ms{$68.5$}{$1.88$} \\
        
        \cmidrule(r){1-1} \cmidrule(lr){2-3} \cmidrule(lr){4-5} \cmidrule(lr){6-7} \cmidrule(l){8-9}
        
        \cellcolor{mypink2}~~w/ DASO~\cite{oh2022daso}
        &\cellcolor{mypink2}\ms{$86.6$}{$0.84$}
        &\cellcolor{mypink2}\ms{$88.8$}{$0.59$}
        &\cellcolor{mypink2}\ms{$71.0$}{$0.95$}
        &\cellcolor{mypink2}\ms{$80.3$}{$0.65$}
        &\cellcolor{mypink2}\ms{$70.0$}{$1.19$}
        &\cellcolor{mypink2}\ms{$78.4$}{$0.80$}
        &\cellcolor{mypink2}\ms{$65.7$}{$1.78$}
        &\cellcolor{mypink2}\ms{$75.3$}{$0.44$} \\

        \cellcolor{mypink}\textbf{~~w/ DASO+IMWA}
        &\cellcolor{mypink}\ms{$\textbf{87.8}$}{$\textbf{0.59}$}
        &\cellcolor{mypink}\ms{$\textbf{89.3}$}{$\textbf{0.84}$}
        &\cellcolor{mypink}\ms{$\textbf{71.7}$}{$\textbf{0.65}$}
        &\cellcolor{mypink}\ms{$\textbf{80.5}$}{$\textbf{0.29}$}
        &\cellcolor{mypink}\ms{$\textbf{70.4}$}{$\textbf{1.33}$}
        &\cellcolor{mypink}\ms{$\textbf{79.1}$}{$\textbf{0.27}$}
        &\cellcolor{mypink}\ms{$\textbf{66.5}$}{$\textbf{2.01}$}
        &\cellcolor{mypink}\ms{$\textbf{76.0}$}{$\textbf{0.96}$} \\
        
        \cellcolor{mypink2}~~w/ ACR~\cite{wei2023towards}
        &\cellcolor{mypink2}\ms{$92.1$}{$0.18$}
        &\cellcolor{mypink2}\ms{$93.5$}{$0.18$}
        &\cellcolor{mypink2}\ms{$85.0$}{$0.09$}
        &\cellcolor{mypink2}\ms{$89.5$}{$0.17$}
        &\cellcolor{mypink2}\ms{$77.1$}{$0.24$}
        &\cellcolor{mypink2}\ms{$83.0$}{$0.32$}
        &\cellcolor{mypink2}\ms{$75.1$}{$0.70$}
        &\cellcolor{mypink2}\ms{$81.5$}{$0.25$} \\

        \cellcolor{mypink}\textbf{~~w/ ACR+IMWA}
        &\cellcolor{mypink}\ms{$\textbf{92.8}$}{$\textbf{0.34}$}
        &\cellcolor{mypink}\ms{$\textbf{94.4}$}{$\textbf{0.14}$}
        &\cellcolor{mypink}\ms{$\textbf{85.5}$}{$\textbf{0.36}$}
        &\cellcolor{mypink}\ms{$\textbf{90.7}$}{$\textbf{0.28}$}
        &\cellcolor{mypink}\ms{$\textbf{78.6}$}{$\textbf{0.40}$}
        &\cellcolor{mypink}\ms{$\textbf{83.9}$}{$\textbf{0.26}$}
        &\cellcolor{mypink}\ms{$\textbf{76.4}$}{$\textbf{0.59}$}
        &\cellcolor{mypink}\ms{$\textbf{81.9}$}{$\textbf{0.12}$} \\
        
        \bottomrule
    \end{tabular}%

\label{tab:CISSIC_2}
\end{table*}

\begin{figure*}[htb]
	\centering  
    \begin{center}
    \includegraphics[width=1.0\linewidth]{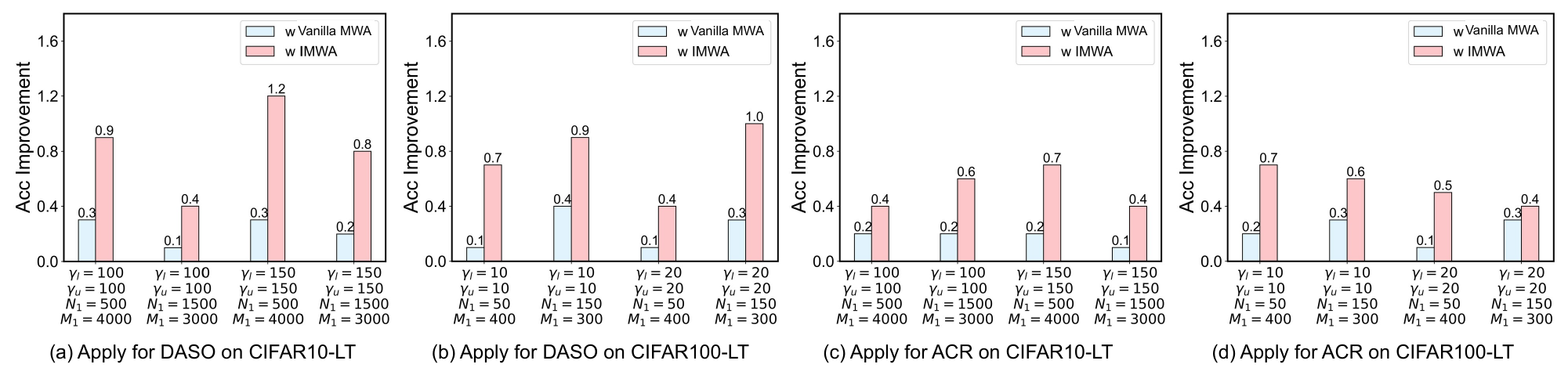}
    \end{center}

	\caption{\textbf{Comparison} of our IMWA and vanilla MWA approaches in terms of achieving performance improvements for DASO~\cite{oh2022daso} and ACR~\cite{wei2023towards}  under consistent class distributions on \textbf{CISSIC benchmarks}, \ie,  \textbf{CIFAR10-LT} and \textbf{CIFAR100-LT}.}

    \label{fig:cissic_mwa_1}
\end{figure*}

\begin{figure*}[htb]
	\centering  
    \begin{center}
    \includegraphics[width=1.0\linewidth]{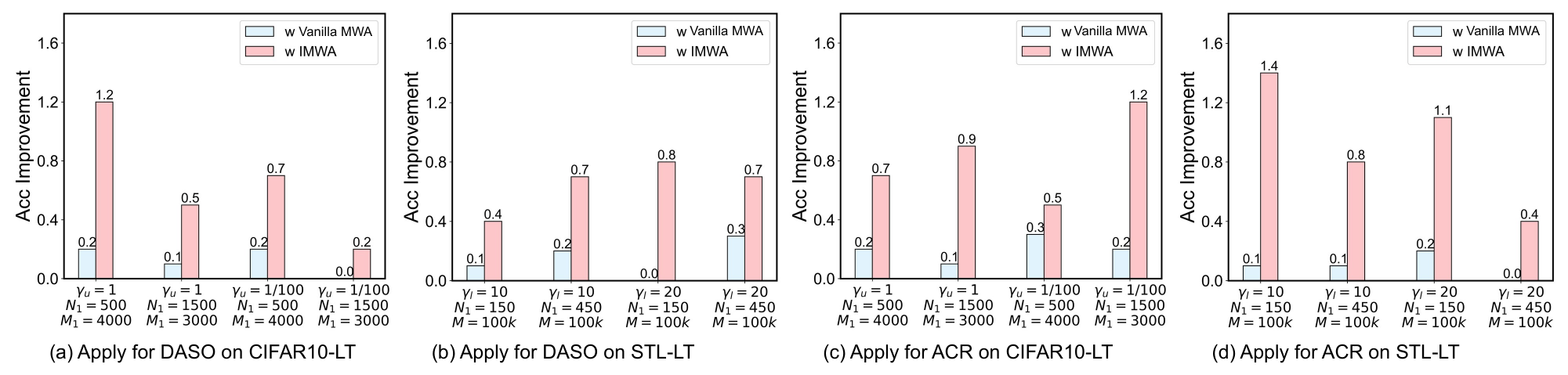}
    \end{center}

	\caption{\textbf{Comparison} of our IMWA and vanilla MWA approaches in terms of achieving performance improvements for DASO~\cite{oh2022daso} and ACR~\cite{wei2023towards}  under inconsistent class distributions on CISSIC benchmarks, \ie, \textbf{CIFAR10-LT} and \textbf{STL-LT}.}

    \label{fig:cissic_mwa_2}
\end{figure*}

\begin{table}[htb]
\centering%
\fontsize{8.3}{11}\selectfont
\centering%
    \caption{\textbf{Performance} of our IMWA applying to DASO~\cite{oh2022daso} and ACR\cite{wei2023towards}  on \textbf{CISSIC benchmark}, \ie, \textbf{ImageNet-127} under different input size. }
    \label{tab:CISSIC_3}
    \centering
    \setlength{\tabcolsep}{18pt}
    \begin{tabular}{l| c  c }

    \toprule
    \textbf{Methods}
    &$\textbf{32} \times \textbf{32}$
    &$\textbf{64} \times \textbf{64}$ \\
    \cmidrule(r){1-1} \cmidrule(lr){2-2} \cmidrule(lr){3-3}
    FixMatch~\cite{sohn2020fixmatch}
    &$29.8$
    &$42.3$ \\
    
    ~~w/ DARP~\cite{kim2020distribution}
    &$30.5$
    &$42.5$\\
    
    ~~w/ DARP+cRT~\cite{kim2020distribution}
    &$39.7$
    &$51.0$\\
    
    ~~w/ CReST+~\cite{wei2021crest}
    &$32.5$
    &$44.7$\\
    
    ~~w/ CReST++LA~\cite{menon2020long}
    &$40.9$
    &$55.9$\\
    
    ~~w/ CoSSL~\cite{fan2022cossl}
    &$43.7$
    &$53.9$\\
    
    ~~w/ TRAS~\cite{wei2022transfer}
    &$46.2$
    &$54.1$\\
    
\cmidrule(r){1-1} \cmidrule(lr){2-2} \cmidrule(lr){3-3}
    
    \cellcolor{mypink2}~~w/ DASO~\cite{oh2022daso}
    &\cellcolor{mypink2}$44.5$
    &\cellcolor{mypink2}$53.7$ \\
    
    \cellcolor{mypink1}~~w/ DASO+vanilla MWA
    &\cellcolor{mypink1}$44.8$
    &\cellcolor{mypink1}$54.3$ \\
    
    \cellcolor{mypink}\textbf{~~w/ DASO+IMWA}
    &\cellcolor{mypink}$\textbf{45.9}$
    &\cellcolor{mypink}$\textbf{55.3}$ \\
    
    \cellcolor{mypink2}~~w/ ACR~\cite{wei2023towards}
    &\cellcolor{mypink2}$57.2$
    &\cellcolor{mypink2}$63.6$ \\
    
    \cellcolor{mypink1}~~w/ ACR+vanilla MWA
    &\cellcolor{mypink1}$57.5$
    &\cellcolor{mypink1}$64.4$ \\
    
     \cellcolor{mypink}\textbf{~~w/ ACR+IMWA}
    &\cellcolor{mypink}$\textbf{59.2}$
    &\cellcolor{mypink}$\textbf{65.7}$ \\
    
    \bottomrule
    \end{tabular}

\end{table}

\begin{table}[htb]
\centering%

\hspace{5mm}%
\centering%
    \caption{\textbf{Performance} of our IMWA applying to Unbiased Teacher~\cite{liu2021unbiased}, Soft Teacher and~\cite{xu2021end} and  Dense Teacher\cite{zhou2022dense}  on \textbf{SSOD benchmark}, \ie, \textbf{MS-COCO}, in terms of $\bm{AP_{50:95}}$. \textit{UBT.} is the abbreviation for Unbiased Teacher; \textit{ST.} is the abbreviation for Soft Teacher; \textit{DT.} is the abbreviation for Dense Teacher.}
    \label{tab:SSOD_1}
    \fontsize{8.3}{11}\selectfont
    \setlength{\tabcolsep}{9.8pt}
    \begin{tabular}{l|c c c c c c c c}
    \toprule
    \textbf{Methods}
    &$\bm {1\%}$
    &$\bm{2\%}$
    &$\bm{5\%}$
    &$\bm{10\%}$  \\ 

    \cmidrule(r){1-1} \cmidrule(lr){2-2} \cmidrule(lr){3-3}  \cmidrule(lr){4-4} \cmidrule(lr){5-5}
    
    STAC~\cite{sohn2020simple}
    &$13.97$
    &$18.25$
    &$24.38$
    &$28.64$ \\
    
    Instant Teaching~\cite{zhou2021instant}
    &$18.05$
    &$22.45$
    &$26.75$
    &$30.40$ \\
    
    ISMT~\cite{yang2021interactive}
    &$18.88$
    &$22.43$
    &$26.37$
    &$30.52$ \\
    
    Humble Teacher~\cite{tang2021humble}
    &$16.96$
    &$21.72$
    &$27.70$
    &$31.61$ \\
    
    Li~\etal~\cite{li2022rethinking}
    &$19.02$
    &$23.34$
    &$28.40$
    &$32.23$ \\
    
    \cmidrule(r){1-1} \cmidrule(lr){2-2} \cmidrule(lr){3-3}  \cmidrule(lr){4-4} \cmidrule(lr){5-5}
    
    \cellcolor{mypink2}\textit{UBT.} \cite{liu2021unbiased}
    &\cellcolor{mypink2}$19.61$
    &\cellcolor{mypink2}$24.13$
    &\cellcolor{mypink2}$27.47$
    &\cellcolor{mypink2}$31.23$  \\ 

    \cellcolor{mypink1}\textit{UBT.}+vanilla MWA
    &\cellcolor{mypink1}$19.97$
    &\cellcolor{mypink1}$24.55$
    &\cellcolor{mypink1}$27.59$
    &\cellcolor{mypink1}$31.64$ \\
    
    \cellcolor{mypink}\textbf{\textit{UBT.}+IMWA}
    &\cellcolor{mypink}$\textbf{20.46}$
    &\cellcolor{mypink}$\textbf{24.56}$
    &\cellcolor{mypink}$\textbf{28.21}$
    &\cellcolor{mypink}$\textbf{32.04}$ \\

    \cellcolor{mypink2}\textit{ST.} \cite{xu2021end}
    &\cellcolor{mypink2}$20.57$
    &\cellcolor{mypink2}$24.85$
    &\cellcolor{mypink2}$30.17$
    &\cellcolor{mypink2}$33.66$  \\  
    
    \cellcolor{mypink1}\textit{ST.}+vanilla MWA
    &\cellcolor{mypink1}$20.74$
    &\cellcolor{mypink1}$25.09$
    &\cellcolor{mypink1}$30.42$
    &\cellcolor{mypink1}$34.01$ \\  
    
    \cellcolor{mypink}\textbf{\textit{ST.}+IMWA}
    &\cellcolor{mypink}$\textbf{21.60}$
    &\cellcolor{mypink}$\textbf{25.49}$
    &\cellcolor{mypink}$\textbf{30.93}$
    &\cellcolor{mypink}$\textbf{34.54}$ \\

    \cellcolor{mypink2}\textit{DT.} \cite{zhou2022dense}
    &\cellcolor{mypink2}$18.33$
    &\cellcolor{mypink2}$25.05$
    &\cellcolor{mypink2}$30.91$
    &\cellcolor{mypink2}$34.51$\\

    \cellcolor{mypink1}\textit{DT.}+vanilla MWA
    &\cellcolor{mypink1}$18.78$
    &\cellcolor{mypink1}$25.32$
    &\cellcolor{mypink1}$31.34$
    &\cellcolor{mypink1}$34.69$ \\

    \cellcolor{mypink}\textbf{\textit{DT.}+IMWA}
    &\cellcolor{mypink}$\textbf{19.06}$
    &\cellcolor{mypink}$\textbf{27.12}$
    &\cellcolor{mypink}$\textbf{32.43}$
    &\cellcolor{mypink}$\textbf{35.87}$ \\
    \bottomrule
    \end{tabular}
    
\end{table}

\begin{table}[htb]
    \centering
    \caption{\textbf{Performance} of our IMWA applying to Unbiased Teacher~\cite{liu2021unbiased}, Soft Teacherand~\cite{xu2021end} and  Dense Teacher\cite{zhou2022dense}  on \textbf{SSOD benchmark}, \ie, \textbf{PASCAL VOC}. \textit{UBT.} is the abbreviation for Unbiased Teacher; \textit{ST.} is the abbreviation for Soft Teacher; \textit{DT.} is the abbreviation for Dense Teacher.}
    \label{tab:SSOD_2}
    \fontsize{8.3}{11}\selectfont
    \setlength{\tabcolsep}{22pt}
    \begin{tabular}{l|c c}
    \toprule
    \textbf{Methods}
    &$\bm{AP_{50}}$
    &$\bm{AP_{50:95}}$ \\
    
    \cmidrule(r){1-1} \cmidrule(lr){2-2} \cmidrule(lr){3-3}
    
    \cellcolor{mypink2}\textit{UBT.} \cite{liu2021unbiased}
    &\cellcolor{mypink2}$81.71$
    &\cellcolor{mypink2}$54.09$ \\

     \cellcolor{mypink1}\textit{UBT.}+vanilla MWA
     &\cellcolor{mypink1}$81.76$
     &\cellcolor{mypink1}$54.13$ \\
     
    \cellcolor{mypink}\textbf{\textit{UBT.}+IMWA }
    &\cellcolor{mypink}$\textbf{82.49}$
    &\cellcolor{mypink}$\textbf{55.88}$ \\
  
    \cellcolor{mypink2}\textit{ST.} \cite{xu2021end}
    &\cellcolor{mypink2}$84.34$
    &\cellcolor{mypink2}$52.51$ \\  
    
    \cellcolor{mypink1}\textit{ST.}+vanilla MWA
    &\cellcolor{mypink1}$84.29$
    &\cellcolor{mypink1}$52.57$ \\  
  
    \cellcolor{mypink}\textbf{\textit{ST.}+IMWA}
    &\cellcolor{mypink}$\textbf{85.17}$
    &\cellcolor{mypink}$\textbf{52.71}$\\
    
    \cellcolor{mypink2}\textit{DT.} \cite{zhou2022dense}
    &\cellcolor{mypink2}$80.27$
    &\cellcolor{mypink2}$56.73$ \\
    
    \cellcolor{mypink1}\textit{DT.}+vanilla MWA
    &\cellcolor{mypink1}$80.29$
    &\cellcolor{mypink1}$56.81$ \\
    
    \cellcolor{mypink}\textbf{\textit{DT.}+IMWA}
    &\cellcolor{mypink}$\textbf{81.15}$
    &\cellcolor{mypink}$\textbf{56.99}$ \\
    \bottomrule
    \end{tabular}
\end{table}

\begin{figure*}[htb]
	\centering  
	\subfigcapskip=-5pt 
	\subfigure[On ImageNet-LT]{
		\includegraphics[width=0.3\linewidth]{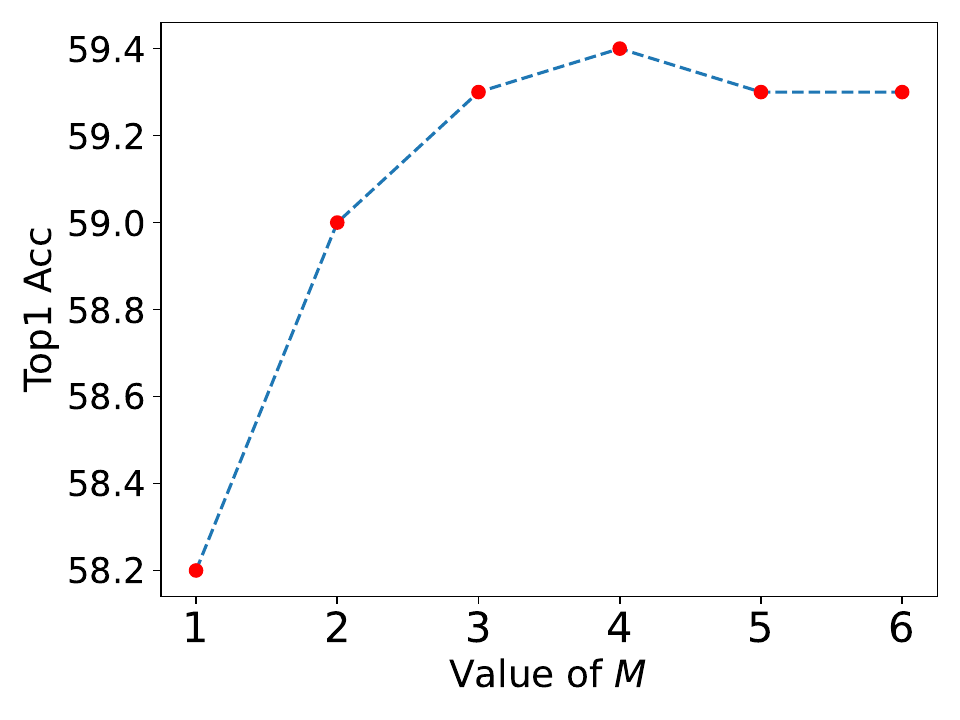}}
	\subfigure[On iNaturalist 2018]{
		\includegraphics[width=0.3\linewidth]{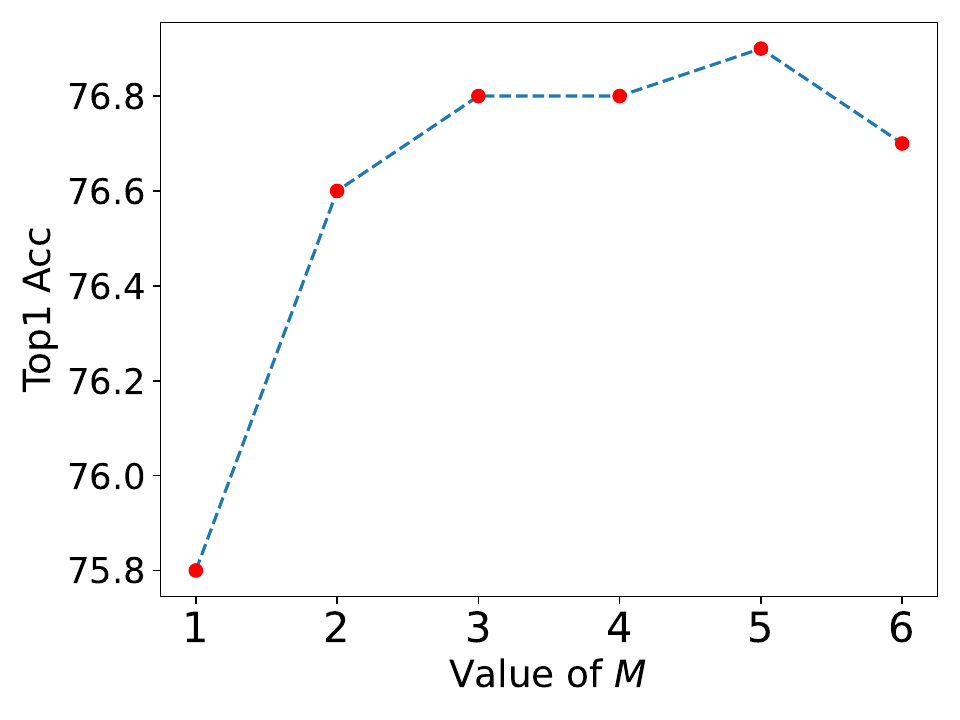}}
	\subfigure[On Places-LT]{
		\includegraphics[width=0.3\linewidth]{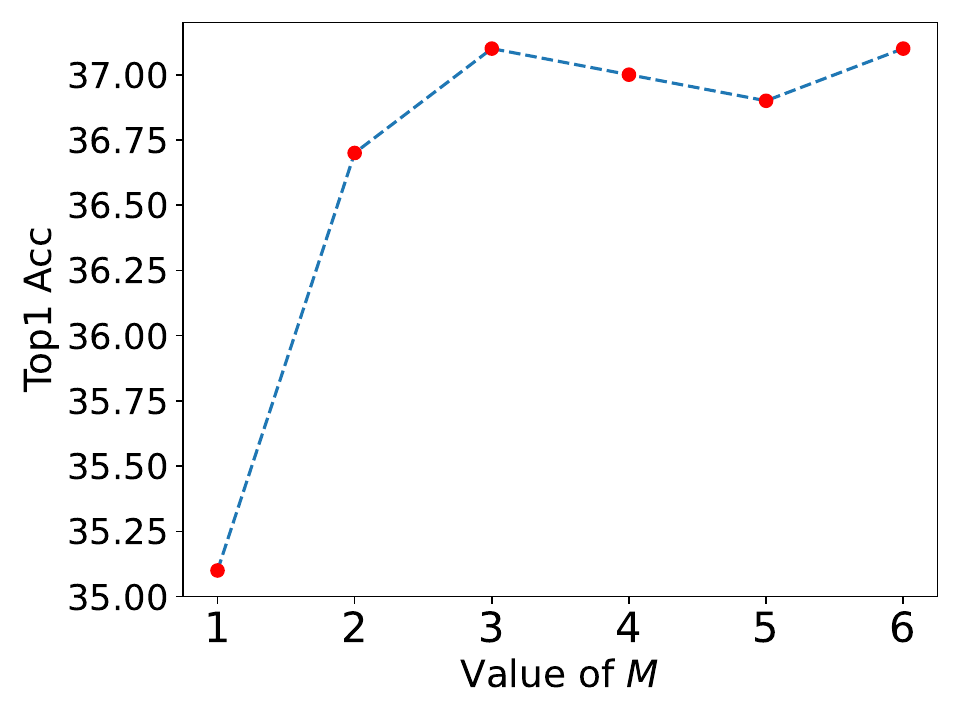}}
     
	\caption{\textbf{Ablation study} of different value of $M$ by LiVT+IMWA on \textbf{three CIIC benchmarks}.}

    \label{fig:model_num}
\end{figure*}

\begin{table}[htb]
\centering
   \caption{\textbf{Ablation study} of different value of $E$. Experiment is conducted by ACR on \textbf{CIIC benchmark}, \ie, \textbf{ImageNet-127} under image size to $32 \times 32$. }
   \label{table:ablation_E}
  \fontsize{8.3}{11}\selectfont
  \setlength{\tabcolsep}{22pt}
   \begin{tabular}{l|c c}
   \toprule
   
   \textbf{Method}
   &$\bm{E}$
   &\textbf{Top1 Acc} \\ 
   
   \cmidrule(r){1-1} \cmidrule(lr){2-2} \cmidrule(lr){3-3}
    \multirow{11}*{ACR+IMWA}
    &\cellcolor{mygray}-
    &\cellcolor{mygray}$57.2$ \\
 
   
    ~ 
    &$1$
    &$57.5$ \\
    
    ~
    &$2$
    &$57.7$ \\
    
    ~
    &$5$
    &$58.7$ \\
    
    ~
    &\cellcolor{mypink2}$\textbf{10}$
    &\cellcolor{mypink2}$\textbf{59.0}$ \\
    
    ~
    &\cellcolor{mypink}$\textbf{20}$
    &\cellcolor{mypink}$\textbf{59.2}$ \\
    
    ~
    &\cellcolor{mypink2}$\textbf{50}$
    &\cellcolor{mypink2}$\textbf{59.0}$ \\
    
    ~
    &\cellcolor{mypink2}$\textbf{100}$
    &\cellcolor{mypink2}$\textbf{59.1}$ \\
    
    ~
    &\cellcolor{mypink2}$\textbf{200}$
    &\cellcolor{mypink2}$\textbf{59.1}$ \\
    
    ~
    &$125000$
    &$57.7$ \\
    
    ~
    &$250000$
    &$57.6$ \\

   \bottomrule
   \end{tabular}
   \end{table}

\subsection{Datasets Details}\label{sec:app_val_datasets}

\subsubsection{Datasets of Class-Imbalanced image classification}
We follow LiVT~\cite{xu2023learning} and BCL~\cite{zhu2022balanced} to perform our IMWA on ImageNet-LT, iNaturalist 2018 and Places-LT benchmarks.
\begin{itemize}
\item {\bf ImageNet-LT: } ImageNet-LT is a long-tailed version ($\gamma=256$) benchmark of vanilla ImageNet~\cite{deng2009imagenet} by sampling a subset following the Pareto distribution with power value $\alpha = 0.6$. 
It consists of 115.8K images of 1000 classes in total with 1280 to 5 images per class.

\item {\bf iNaturalist 2018: } iNaturalist 2018~\cite{van2018inaturalist} is a large-scale dataset containing 437.5K images from 8,142 classes over multiple natural species and suffers from extremely class-imbalanced distribution ($\gamma=512$).

\item {\bf Places-LT: } Places-LT is a long-tailed variant of the large-scale scene classification dataset Places~\cite{zhou2017places}, which contains 62.5K images from 365 categories of scenes in total with 4,980 to 5 ($\gamma=996$).

\end{itemize}

\subsubsection{Datasets of Class-Imbalanced Semi-Supervised Image Classification} 
We follow ACR~\cite{wei2023towards} and DASO~\cite{oh2022daso} to perform our IMWA on CIFAR-10/100-LT~\cite{krizhevsky2009learning}, STL-10-LT~\cite{coates2011analysis} and ImageNet-127~\cite{fan2022cossl}. Given a imbalance ratio $\gamma^l$ of labeled set, we set the number of labeled samples for class $c$ as $N_c$; given the imbalance ratio of unlabeled set $\gamma^u$, we set the number of labeled samples for class $c$ as $N_c = $
for class c as $M_c = M_1 \cdot {\gamma^u}^{-\frac{c-1}{C-1}}$ with given $M_1$.
\begin{itemize}
	\item {\bf CIFAR-10-LT: } Following ACR~\cite{wei2023towards} and   and DASO~\cite{oh2022daso}, we perform our IMWA under $N_1 = 500, M_1 = 4000$ and $N_1 = 1500, M_1 = 3000$. Meanwhile, we set $\gamma^l=\gamma^u=100$ and $\gamma^l=\gamma^u=150$ respectively. We also perform IMWA on \textit{uniform} ($\gamma^l=100$) and \textit{reversed} ($\gamma^u\ \in \{1,1/100\}$) mentioned in DASO~\cite{oh2022daso} to simulate various class distribution of unlabeled data.

	\item {\bf  CIFAR-100-LT: } We test our IMWA under $N_1 = 50, M_1 = 400$ and $N_1 = 150, M_1 = 300$. The imbalance ratio is set to $\gamma^l = \gamma^u = 10$ and $\gamma^l = \gamma^u = 20$. For the uniform and reversed unlabeled data class distributions, we set $\gamma^l = 10$ and $\gamma^u \in \{1, 1/10\}$.

	\item {\bf  STL-10-LT: } Following DASO, we test our IMWA under $\gamma^l \in \{10,20\}$. Note that the ground-truths  of unlabeled set are unknown, therefore we simply use all the unlabeled data instead of sampling.

	\item {\bf  ImageNet-127: } ImageNet127 is a naturally class-imbalanced dataset in long-tailed distribution, hence we do not need to construct the datasets manually.
 Following ACR, we downsample the image size to 32 × 32 and 64 × 64.

\end{itemize}

\subsubsection{Datasets of  Semi-Supervised Object Detection} 
We follow Unbiased Teacher~\cite{liu2021unbiased}, Soft Teacher~\cite{xu2021end} and Dense Teacher~\cite{zhou2022dense} to perform our IMWA on PASCAL VOC ~\cite{everingham2010pascal} and MS-COCO~\cite{lin2014microsoft} benchmarks.
\begin{itemize}
	\item {\bf MS-COCO: } MS-COCO~\cite{lin2014microsoft} dataset is a large scale visual recognition benchmark including 80 object categories  with a total of 123k labeled images for object detection task, where the \texttt{train2017} set and \texttt{val2017} set contain 118k images and 5k images respectively. To build the semi-supervised setting, we follow the data split protocol of~\cite{liu2021unbiased} to sample 1\%, 2\%, 5\% and 10\% images of the \texttt{train2017} set as the labeled set. And the rest part of images are treated as the unlabeled set. \texttt{val2017} is viewed as the test set to evaluate our method.

	\item {\bf PASCAL VOC: } PASCAL VOC 2007 and 2012 datasets \cite{everingham2010pascal} contain 9,963 labeled images and 22,531 labeled images respectively including 20 categories. Following \cite{liu2021unbiased,xu2021end,zhou2022dense}, we
treat the \texttt{VOC 07 trainval set} as the labeled set and the \texttt{VOC 12 trainval set} as the unlabeled set. \texttt{VOC 07 test set} is viewed as the test set to evaluate our method.

\end{itemize}

\subsection{Main Results}
Here we report the experimental results for three class-imbalanced learning tasks.
\subsubsection{\textcolor{black}{Comparison on Class-Imbalanced Image Classification (CIIC)}}
Tab.~\ref{tab:compare_w_sota_ciic} shows the comparison results with previous state-of-the-art (SoTA) methods, 
where ``+IMWA'' means performing experiments with our IMWA.
According to Tab.~\ref{tab:compare_w_sota_ciic}, BCL+IMWA outperforms BCL \textbf{1.3\%}, \textbf{0.7\%} and \textbf{0.9\%} in terms of Top-1 Acc on ImageNet-LT, iNaturalist 2018 and Places-LT respectively. 
LiVT+IMWA outperforms LiVT \textbf{0.8\%}, \textbf{0.8\%} and \textbf{1.6\%} in terms of Top-1 Acc on ImageNet-LT, iNaturalist 2018 and Places-LT respectively. 
Especially, LiVT+IMWA achieves the state-of-the-art performance on ImageNet-LT and iNaturalist 2018 benchmarks.
Note that BCL employs a CNN backbone and LiVT employs a ViT~\cite{dosovitskiy2020image} backbone, which implies that our IMWA is effective for different types of backbones. 
Fig.~\ref{fig:ciic_ema} shows the comparison results with vanilla MWA and EMA for CIIC task on three benchmarks.
\textcolor{black}{For comparison on ImageNet, BCL and LiVT with vanilla MWA only improve about 0.2\% in terms of Top1-Acc, while IMWA obtains \textbf{1.8\%} and \textbf{1.4\%} improvements respectively. \textit{Notice that  vanilla MWA has the same total training iterations
with IMWA}, thus our IMWA can achieve higher improvement than vanilla MWA does without extra computation cost.}
%
 We also observe that our IMWA shows comparable improvements with EMA (1.3\% vs. 1.1\% for BCL, 0.8\% vs. 1.1\% for LiVT),
while performing both IMWA with EMA results in further  performance improvement (\textbf{1.8\%} for BCL and \textbf{1.4\%} for LiVT), which implies that EMA and our IMWA complement each other.
We can also observe the similar comparison results on iNat-18 and Places-LT datasets.
\subsubsection{\textcolor{black}{Comparison on Class-Imbalanced Semi-Supervised Image Classification (CISSIC) and on Semi-Supervised Object Detection (SSOD)}}
Similar to CIIC, we apply IMWA to the CISSIC and SSOD tasks.
%
Tab.~\ref{tab:CISSIC_1},Tab.~\ref{tab:CISSIC_2},Tab.~\ref{tab:CISSIC_3}, Fig.~\ref{fig:cissic_mwa_1} and ~\ref{fig:cissic_mwa_2}
 show the extensive experimental comparison results in terms of Top-1 Acc on CISSIC tasks; 
%
Tab.~\ref{tab:SSOD_1} to Tab.~\ref{tab:SSOD_2} show the extensive experimental comparison results in terms of $AP_{50:95}$ and $AP_{50}$ on SSOD tasks.
Let's take the experimental results on the SSOD task as an example.
According to these results, we analyze the effectiveness of our IMWA over the baseline methods in four-fold as follows:

\begin{itemize}
  \item [1)] 
   Our IMWA achieves \textcolor{black}{consistent} 
   improvements for different proportion of labeled set. Let's take the results of Dense Teacher for an example, as shown in Tab.~\ref{tab:SSOD_1}. 
   \textcolor{black}{Over each proportion of labeled setting (1\%,2\%,5\% and 10\% labeled ratio), Dense Teacher + IMWA achieve \textbf{0.73\%} to \textbf{2.52\%} $AP_{50:95}$ improvements consistently.}
  \item [2)]
   Our method achieves stable improvements for various baseline methods. As shown in Tab.~\ref{tab:SSOD_2}, with respect to their corresponding baseline methods, our methods obtain \textbf{0.73\%} to \textbf{1.03\%} $AP_{50:95}$ improvements on 1\% labeled images, \textbf{0.43\%} to \textbf{2.07\%} $AP_{50:95}$ on 2\% labeled images, \textbf{0.74\%} to \textbf{2.52\%} $AP_{50:95}$ on 5\% labeled images, \textbf{0.81\%} to \textbf{1.36\%} $AP_{50:95}$ on 10\% labeled images.
  \item [3)] 
   Our method achieves stable improvements for different datasets. Tab.~\ref{tab:SSOD_2} shows the experimental results on PASCAL VOC benchmark. Our methods based on three baseline achieve similar margin of improvements with that of MS-COCO.
    \item [4)] 
   Our method achieves stable outperforming comparing with the vanilla MWA. Extensive comparison from Tab.~\ref{tab:SSOD_1} and Tab.~\ref{tab:SSOD_2} show that  the improvement brought by vanilla MWA is significantly lower than that of our IMWA, which indicates that our IMWA is superior to vanilla MWA.
\end{itemize}
\noindent Similar conclusions can be drawn from the experimental results on CISSIC in Tab.~\ref{tab:CISSIC_1},~\ref{tab:CISSIC_2},~\ref{tab:CISSIC_3} and Fig.~\ref{fig:cissic_mwa_1},~\ref{fig:cissic_mwa_2} as well.
Based on the above analysis, our IMWA can bring stable improvement over various class-imbalanced learning tasks and evaluation benchmark. 
Moreover, from the point of view of implementation, IMWA can be easily adapted into any existing CISSIC and SSOD without modifying models' structure or adding complicated tricks, which shows the sufficient efficiency of IMWA.

\subsection{Ablation Studies}
In this section we conduct ablation studies to explore the effect of $M$ and $E$, which are the hyper-parameters of our IMWA.
\subsubsection{Effect of the number of models $M$} $M$ is a hyper-parameters of our IMWA, which presents the number of models involving training. 
Intuitively, the larger the value of $M$, the better the IMWA effect.
The existing MWA appoarch~\cite{wortsman2022model,rame2022diverse} also propose that increasing the number of models can enhance performance.
However, as M increases, the computational cost of the whole training process also increases. 
Therefore, it is essential to search a suitable $M$ to balance the performance and computational cost.
Here we conduct ablation study by tuning the  $M$ from $2$ to $6$ for LiVT+IMWA on three benchmarks, and the results are shown in Fig.~\ref{fig:model_num}. 
$M=1$ represents the setting without our IMWA.
When $M=2$ or $M=3$, the performance of IMWA are increase obviously.  
As the $M$ proceed to increase, there are no significant improvement in terms of performance. Note that a large $M$ will aggravate the both cost of computation and memory. Thus, we suppose $\bm{M=2}$ or $\bm{M=3}$ are  trade-off decisions in implementation between performance and efficiency.

\subsubsection{Effect of the number of episode $E$} $E$ represents the number of episodes in our IMWA. 
Larger value of $E$ means that the average operation between models will be more frequent.
Intuitively, the more frequent the average operation, the higher the performance will be obtained, as each average operation could potentially lead to a performance improvement. 
However, when $E$ increases, the training iterations of each episode will be reduced, which will lead to the low diversity among the models to be averaged. 
Existing work~\cite{rame2022diverse} has proved that the effect of MWA is positively correlated with the differences between models.
Therefore, we conduct ablation study to explore the best choice of $E$ to balance the average frequency and models' diversity.
We perform ACR+IMWA on ImageNet-127 under $E$ from $1$ to $250000$, where $250000$ is the default total training iteration of ACR~\cite{wei2023towards}.
Tab.~\ref{table:ablation_E} shows the ablation results of various $E$.
The first row represent the ACR w/o IMWA setting and the second row ($E=1$) represents the vanilla MWA approach.
The experimental results show that too large or too small $E$ lead to slight improvement compared with baseline (w/o IMWA setting).
While $E$ is in the range of \textbf{5} to \textbf{200}, the improvement of performance is more noticeable.
The above results indicate that there is relatively flexible room for tunning $E$, only necessary to avoid extremely large or extremely small values.

\subsection{Discussion}\label{sec:app_discussion}
In this section, we further give some discussion about our IMWA.

\begin{table}[t]
   \caption{\textbf{GPU memory cost} for Unbiased Teacher+IMWA under different $M$ on \textbf{SSOD benchmark}, \ie, \textbf{PASCAL VOC}. }
   \label{table:memory}
   \begin{center}
    \fontsize{8.3}{11}\selectfont
    \setlength{\tabcolsep}{10.2pt}
   \begin{tabular}{l|c c}
   \toprule
   {\bf Method} &{$\bm{M}$}&{\textbf{GPU memory cost / GB} }  \\ 
   \cmidrule(r){1-1} \cmidrule(lr){2-2} \cmidrule(lr){3-3}
   \multirow{6}*{Unbiased Teacher+IMWA}
   &\cellcolor{mygray}$1$
   &\cellcolor{mygray}$6.63$ \\
   
    ~
    &$2$
    &$7.35$ \\
    
    ~
    &$3$
    &$8.06$ \\
    
    ~
    &$4$
    &$8.76$ \\
    
    ~
    &$5$
    &$9.45$ \\
    
    ~
    &$6$
    &$10.47$\\

   \bottomrule
   \end{tabular}
   \end{center}
   \end{table}

\subsubsection{Discussion about Memory Consumption}
Due to training multiple models simultaneously in IMWA, it inevitably introduces additional computational overhead like vanilla MWA.
To save the GPU memory cost during training, in our implementation of IMWA, we equivalently transfer the “parallel training” into a “serial training” manner to avoid the large cost of GPU memory. 
Specifically, in one episode, we can train each model $\bm{f}$ sequentially, and then save its trained weights into a dictionary (a data structure of python). 
After all the M models are well trained in this episode, we average their weights and then start the next episode. 
By this way, the GPU memory cost from simultaneous forward and backward propagation of M pairs detectors can be avoided, while only the trained weights need to be stored in GPU memory. 
We take Unbiased Teacher+IMWA under different model number $M$ as an example.
Tabel.~\ref{table:memory} shows the results of actual GPU memory cost.
For adding more one pair of model (optimized model and EMA model), the corresponding memory cost will increase by $(10.47-6.63)/5=0.76$ GB on average for Unbiased Teacher, \textit{which are acceptable GPU memory costs for training them on each RTX 2080Ti (about 11 GB memory)}.

\begin{table}[t]
    \centering
    \caption{\textbf{Comparison} of IMWA with baseline methods trained with $2 \times T$ iterations on \textbf{SSOD benchmark}, \ie, \textbf{PASCAL VOC}. Experiments are conducted by Unbiased Teacher, Soft Teacher and Dense Teacher.}

    \label{tab:3x}
    \fontsize{8.3}{11}\selectfont
    \setlength{\tabcolsep}{12.8pt}
    \begin{tabular}{l|c  c | c}
    \toprule
    \bf{Methods}  &  \bm{$2 \times$} & \bf IMWA & $\bm{AP_{50}}$ \\
    \cmidrule(r){1-1} \cmidrule(r){2-3} \cmidrule(r){4-4}
    
    \multirow{3}*{Unbiasd Teacher}
    &
    &
    &$81.71$  \\
    
    ~
    &\cellcolor{mypink2}\checkmark
    &\cellcolor{mypink2}
    &\cellcolor{mypink2}$81.74 (+0.03)$ \\
    
    ~
    &\cellcolor{mypink}
    &\cellcolor{mypink}\checkmark
    &\cellcolor{mypink}$\textbf{82.49 (+0.78)}$\\ 
    
    \cmidrule(r){1-1} \cmidrule(r){2-3} \cmidrule(r){4-4}

    \multirow{3}*{Soft Teacher}
    &
    &
    &$84.34$ \\
    
    ~
    &\cellcolor{mypink2}\checkmark
    &\cellcolor{mypink2}
    &\cellcolor{mypink2}$84.41 (+0.07)$ \\

    ~
    &\cellcolor{mypink}
    &\cellcolor{mypink}\checkmark
    &\cellcolor{mypink}$\textbf{85.17 (+0.83)}$\\

    \cmidrule(r){1-1} \cmidrule(r){2-3} \cmidrule(r){4-4}

    \multirow{3}*{Dense Teacher}
    &
    &
    &$80.27$ \\
    
    ~
    &\cellcolor{mypink2}\checkmark
    &\cellcolor{mypink2}
    &\cellcolor{mypink2}$80.32 (+0.05)$\\

    ~
    &\cellcolor{mypink}
    &\cellcolor{mypink}\checkmark
    &\cellcolor{mypink}$\textbf{81.15 (+0.88)}$ \\
    \bottomrule
    \end{tabular}
\end{table}

\subsubsection{Comparison with Baseline under Longer Training Schedule}
In our IMWA, $M$ models are trained in parallel, each of which is trained in total of $T$ iterations.
Therefore, the whole training iterations of IMWA is $M \times T$, which is about $M$ times more than that of baseline method (only $T$ iterations). 
For a fair comparison, we train the  baseline methods with the same training iterations as IMWA (\eg $M \times T$ iterations.
Here we choose to conduct experiments on SSOD task as an example.
Specifically, we train Unbiased Teacher, Soft Teacher and Dense Teacher for $M \times T$ iterations respectively, and compare with w/ IMWA under $M=2$ for fair comparison.
 Table~\ref{tab:3x} shows the experimental results. Simply training the baseline method  with $M \times$ iterations shows less improvements, while our IMWA obtains more obvious improvements. \textit{From this point of view, training a SSOD method in IMWA pipeline will benefit more than simply training baseline method with more iterations}.

\begin{table}[htb]
    \centering
    \caption{\textbf{Impact} of dataset imbalance ratio $r$ for IMWA. Experiments are conducted by WideResNet-28 using vanilla supervised manner on \textbf{CIIC benchmark}, \ie, \textbf{CIFAR-10-LT}.}

    \label{tab:im}
    \fontsize{8.3}{11}\selectfont
    \setlength{\tabcolsep}{17.8pt}
    \begin{tabular}{l|c  c }
    \toprule
    \bf{Imbalance ratio}  &  \bf{w/o IMWA} & \bf{w/ IMWA} \\
    \cmidrule(r){1-1} \cmidrule(r){2-2} \cmidrule(r){3-3}
    
    \cellcolor{mygray}$1$
    &\cellcolor{mygray}$84.2$
    &\cellcolor{mygray}$84.4~(+0.2)$ \\

    $5$
    &$72.5$
    &$73.4~(+0.9)$ \\
    
    $10$
    &$61.5$
    &$63.8~(+\textbf{2.3})$\\

    $20$
    &$53.5$
    &$54.5~(+1.0)$ \\
 
    $50$
    &$42.5$
    &$44.0~(+1.5)$\\ 
    
    $100$
    &$36.8$
    &$37.8~(+1.0)$\\

    $200$
    &$33.1$
    &$33.4~(+0.3)$\\

    \bottomrule
    \end{tabular}
\end{table}

\subsubsection{Discussion about the imbalance ratio}
In this section we aim to exploring how the imbalance ratio of training set impact the effectiveness of our IMWA.
We conduct experiments on CIFAR-10-LT with WideResNet-28 for CIIC task under various imbalance ratio $\gamma \in \{1,5, 10,20,50,100,200\}$.
In particular, the experiment with $\gamma =1$ is equivalent to a balanced setting.
From the Tab.~\ref{tab:im}, we can observe that the performance improvement are not significant when the $\gamma$ is too small or too large, which suggests that our IMWA is better suited for scenarios with class imbalance but not severe.
For example, when $\gamma =10$, IMWA brings absolutely \textbf{2.3\%} improvement in terms of Top 1 Accuracy.
We will explore how to enhance the performance of IMWA in scenarios with either class balance or severe class imbalance in our future work. 

\subsubsection{Comparison with concurrent works}
We have surveyed two very recent work~\cite{zhang2023lookaround,jain2023dart} whose idea are  similar with our IMWA. 
In general, there are three differences between our IMWA and these works. 
First, we evaluate the effectiveness of IMWA on various class-imbalanced learning task, with analyzing the reason of performing better in a class-imbalanced setting. While these concurrent only evaluate their methods on standard image classification task or out-of-distribution scenario.
Second, we further explore the synergistic collaboration of our IMWA on EMA-based methods and empirically demonstrate that IMWA and EMA can be mutual promotion, which is something that these concurrent works have not done.
Third, in our implementation, the improvement can be obtained without adopting different data augmentation or enumerate various hyper-parameters (\eg learning rate) for each individual model, which these concurrent works need.
In summary, both IMWA and these concurrent works demonstrate the effectiveness of MWA with an ``iterative'' manner.
However, in our work, we further illustrate that IMWA can be applied to a broader range of scenarios and mutual promotion with other MWA technique.

\section{Conclusion}

In this paper, we empirically explore the current model weight averaging (MWA). 
Specifically, the vanilla MWA can benefit the class-imbalanced learning, and performing model averaging in the early epochs of training yields a greater performance improvement than doing that in later epochs.
\textcolor{black}{Based on this discovery, we  propose a novel MWA technique named IMWA for class-imbalanced learning tasks, which present the vanilla MWA for multiple times during the training process to further obtain performance improvement.}
%
\textcolor{black}{We apply IMWA on several existing SOTA methods of three class-imbalanced image recognize tasks to evaluate its effectiveness. In addition, we also explore the synergistic collaboration of IMWA and EMA. Extensive experiments results on various benchmarks demonstrate that IMWA can achieve larger performance improvements for these SOTA methods than vanilla MWA does, while IMWA and
EMA can complement each other.}

\bibliographystyle{ieee_fullname}
\bibliography{reference}











\newpage

\section{Biography Section}

\begin{IEEEbiography}[{\includegraphics[width=1in,height=1.25in,clip,keepaspectratio]{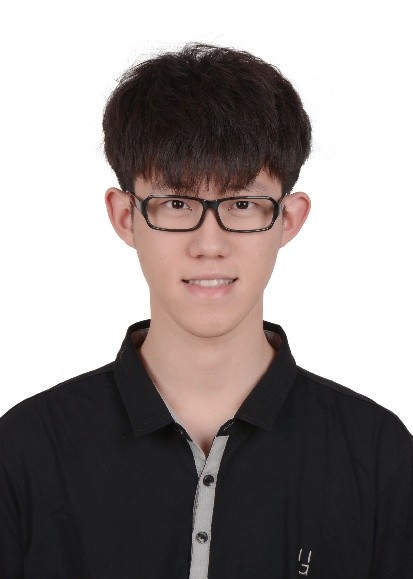}}]{Zitong Huang} is currently pursuing the Ph.D. degree with Harbin Institute of Technology, Harbin, China. His research interests include computer vision, deep learning, continual learning and object detection.
\end{IEEEbiography}

\begin{IEEEbiography}[{\includegraphics[width=1in,height=1.25in,clip,keepaspectratio]{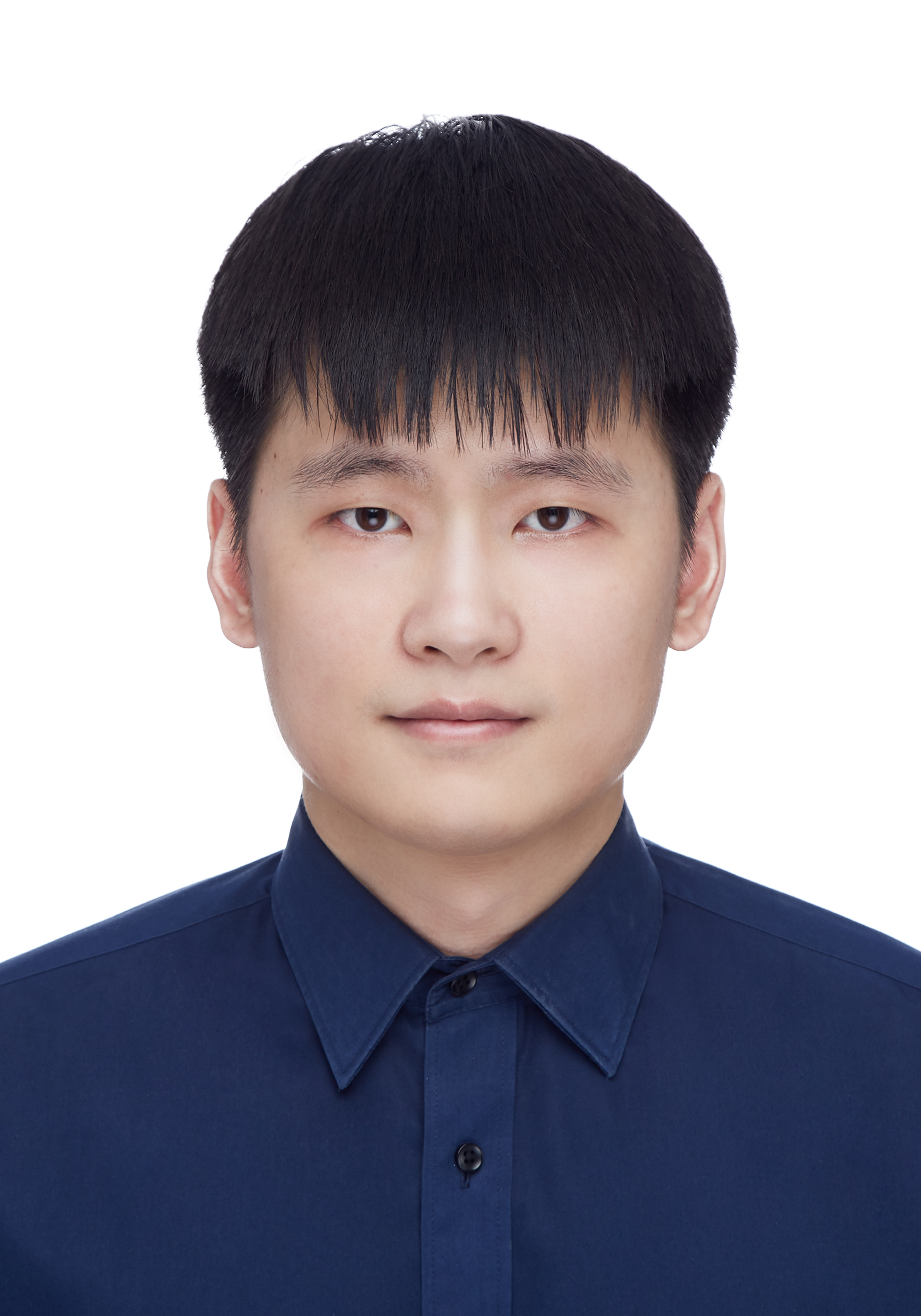}}]{Ze chen} is currently employed at Megvii Technology Limited, having graduated with a master's degree from Shanghai Jiao Tong University. His research interests encompass general object detection, the practical applications of deep learning models, and generative modeling.
\end{IEEEbiography}

\begin{IEEEbiography}[{\includegraphics[width=1in,height=1.25in,clip,keepaspectratio]{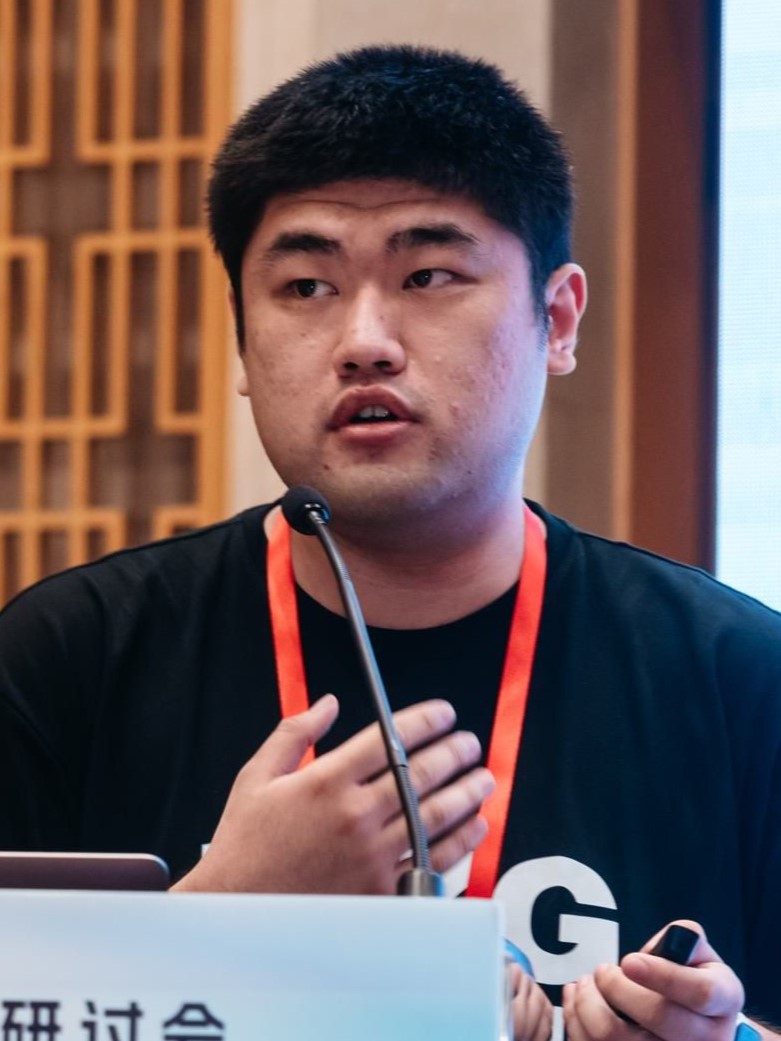}}]{Bowen Dong} is the PhD Candidate of Harbin Institute of Technology and Hong Kong Polytechnic University, and his supervisors are Professor Wangmeng Zuo and Professor Lei Zhang. His main research directions are weakly supervised learning, few-shot learning and transfer learning. He has published several papers in top conferences such as ICCV, ECCV, CVPR, and ICLR, and has served as a reviewer for conferences or journals such as CVPR, ICCV, ECCV, and TIP.
\end{IEEEbiography}

\begin{IEEEbiography}[{\includegraphics[width=1in,height=1.25in,clip,keepaspectratio]{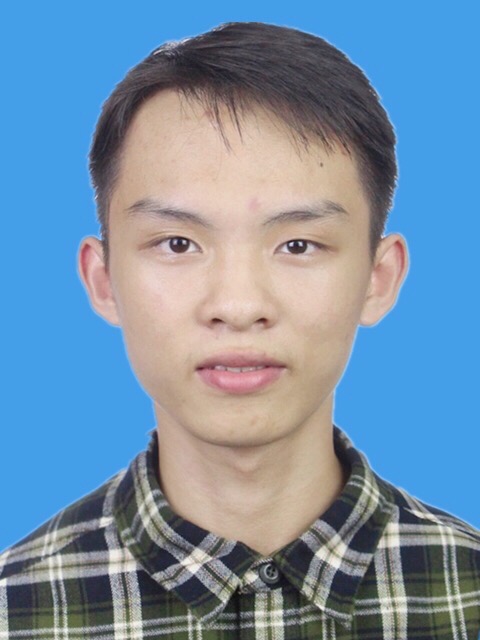}}]{Chaoqi Liang}  is currently pursuing a Bachelor of Science degree at Harbin Institute of Technology in Harbin, China. His research interests include computer vision, large language models, semi-supervised learning, and computational biology.
\end{IEEEbiography}

\begin{IEEEbiography}[{\includegraphics[width=1in,height=1.25in,clip,keepaspectratio]{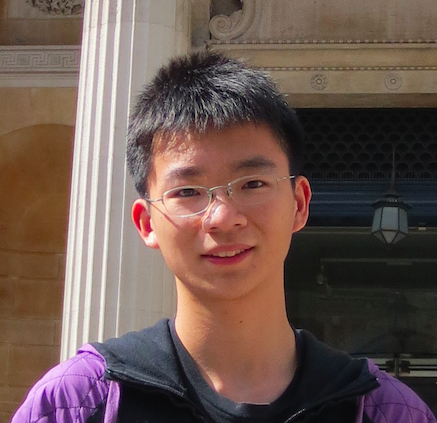}}]{Erjin Zhou} is the Director of the research group at Megvii Research Institute, responsible for building and leading the team to conduct research on facial and human body detection and recognition, portrait generation, general object detection, and action recognition technologies. Erjin Zhou also leads the team in the research and development of algorithm production tools. His research achievements have been applied to Megvii's cloud-based identity verification solution, as well as industry solutions for intelligent building access, smart phone AI solutions, and financial industry identity verification. 
\end{IEEEbiography}

\begin{IEEEbiography}
[{\includegraphics[width=1in,height=1.25in,clip,keepaspectratio]{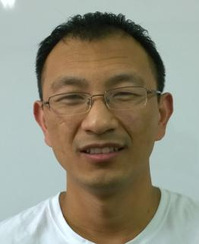}}]{Wangmeng Zuo} received the Ph.D. degree from the Harbin Institute of Technology in 2007. He is currently a Professor in the School of Computer Science and Technology, Harbin Institute of Technology. His research interests include image enhancement and restoration, image and face editing, object detection, visual tracking, and image classification. He has published over 100 papers in top tier journals and conferences. His publications have been cited more than 55,000 times. He also serves as Associate Editors for IEEE T-PAMI and IEEE T-IP.
\end{IEEEbiography}
 




\vfill

\end{document}